\documentclass[manuscript,screen]{acmart}
\usepackage{booktabs}

\AtBeginDocument{%
  }

\setcopyright{acmlicensed}
\copyrightyear{2026}
\acmYear{2026}
\acmDOI{XXXXXXX.XXXXXXX}
\acmJournal{TALLIP}

\begin{document}

\title[Gaokerena]{Gaokerena: A Small Persian Medical Language Model Family}

\author{Mehrdad Ghassabi}
\affiliation{\institution{University of Isfahan}\city{Isfahan}\country{Iran}}
\author{Hamidreza Baradaran Kashani}
\affiliation{\institution{University of Isfahan}\city{Isfahan}\country{Iran}}
\author{Pedram Rostami}
\affiliation{\institution{University of Tehran}\city{Tehran}\country{Iran}}
\author{Sadra Hakim}
\affiliation{\institution{University of Windsor}\city{Windsor}\state{ON}\country{Canada}}
\author{Zahra Kazemi}
\affiliation{\institution{Alzahra University}\city{Tehran}\country{Iran}}
\author{Amirhossein Poursina}
\affiliation{\institution{Isfahan University of Medical Sciences}\city{Isfahan}\country{Iran}}
\author{Milad Tavakoli}
\affiliation{\institution{University of Isfahan}\city{Isfahan}\country{Iran}}
\author{Audrina Ebrahimi}
\authornote{Corresponding author.}
\email{audrina.ebrahimi@utdallas.edu}
\affiliation{\institution{University of Texas at Dallas}\city{Dallas}\state{TX}\country{USA}}
\renewcommand{\shortauthors}{Ghassabi et al.}

\begin{abstract}
Persian medical natural language processing remains limited by the scarcity of domain-specific resources and by insufficient public characterization of their coverage, quality, and limitations. We present Gaokerena, a suite of Persian medical resources, comprising an 85.5-million-word corpus of 101,954 web articles; MF3QA, a 24,000-pair patient--doctor question--answering dataset selected from a raw crawl of 186,029 pairs and supplemented with 201 translated K-QA items; 1,878 Iranian medical-exam questions (KOPP); 1,089 manually translated medical MMLU questions (FA\_MED\_MMLU); and approximately 18,000 machine-translated MedMCQA training items filtered by two language-model evaluators. We characterize source composition, orthographic variation, redundancy, topical coverage, QA selection effects, privacy indicators, answer-key balance, extraction artefacts, and translation limitations. As application case studies, we adapt Aya-Expanse-8B in two ways: Gaokerena-V, trained on our corpus and MF3QA, and Gaokerena-R, trained with preference-based reasoning distillation to improve medical reasoning. On translated medical MMLU, Gaokerena-V achieves 49.31\% accuracy under direct prompting, compared with 46.65\% for Aya-Expanse-8B; a paired exact McNemar test yields $p=0.0502$, which is statistically inconclusive. Gaokerena-R reaches 52.80\% with five-sample self-consistency and a verifier, although this higher-compute configuration is not matched to the directly prompted backbone. Under matched five-sample chain-of-thought self-consistency without a verifier, Gaokerena-R scores 48.58\% (529/1,089), versus 46.92\% (511/1,089) for Aya-Expanse-8B; the paired exact McNemar test is not significant ($p=0.190$). English evaluation also reveals a substantial performance gap relative to Persian. To our knowledge, this work provides the first openly released, characterized suite of Persian medical resources together with generative models built on them; the model results are protocol-specific case studies and do not establish an isolated training advantage or clinical readiness.
\end{abstract}

\maketitle

\section{Introduction}\label{sec1}

Persian medical natural language processing (NLP) faces two related constraints: a limited supply of domain-specific resources and limited public evidence about the coverage, quality, and suitability of the resources that do exist. Medical NLP depends not only on model architectures but also on the data used to train, adapt, and evaluate them. Relevant Persian material is distributed across health websites, patient--doctor forums, translated medical question sets, and national examination papers. These sources differ in genre, editorial practice, orthographic conventions, reliability, and intended use, and should not be treated as interchangeable or uniformly representative.

Compact language models are one possible way to make domain-specific NLP more accessible. Compared with remote general-purpose systems, models that can run locally may offer benefits in cost, latency, connectivity, and data governance. Those benefits depend on the deployment setting, however, and model size alone does not establish medical competence or safety. For medical applications, resource provenance, data quality, evaluation design, and limitations are as important as headline accuracy.

Persian language technology has benefited from multilingual models such as Aya-Expanse~\cite{b12,b13} and Persian--English models such as PersianMind~\cite{b14}; the Persian medical domain has also been addressed by resources such as Sina-BERT~\cite{b6}. These efforts do not remove the need for accessible generative models and domain-specific datasets whose construction and limitations can be inspected. Public medical text and patient--doctor exchanges can contain repeated material, uneven answer quality, orthographic variation, sensitive information, and strong source-specific biases. Benchmark questions may also carry extraction errors or translation artefacts. Characterizing these properties is necessary for researchers to interpret model performance and to decide whether a dataset is suitable for a particular use.

This article presents Gaokerena as a suite of Persian medical language resources and uses two small models, Gaokerena-V and Gaokerena-R, to illustrate how the resources can support domain adaptation, instruction tuning, and reasoning-oriented post-training. The central contribution is the resource suite and its characterization: a medical web corpus, a patient--doctor QA dataset, a structured Iranian medical-exam collection, a translated medical benchmark, and a translated MCQ training resource. The model experiments show what these resources can support under particular evaluation protocols; they do not isolate the causal contribution of each resource or establish a general training advantage. We also compare Aya-Expanse-8B on corresponding English and Persian medical MMLU questions to describe the observed language gap under the protocols used here.

\paragraph{Relation to earlier publications.} This article extends our conference papers on Gaokerena-V~\cite{b1} and Gaokerena-R~\cite{b2}, and summarizes the separately reported claim-level uncertainty heads~\cite{b3}. The earlier papers introduced the models and reported initial experiments. Audrina Ebrahimi is a co-author of this expanded journal article but was not an author of the two earlier conference papers; her contribution applies to the present extended study and is specified in the Author Contributions statement. This article adds systematic analyses of corpus composition, orthography, redundancy, boilerplate, and topic coverage; analyses of MF3QA source confounding, selection effects, alignment, repetition, overlap, privacy-related indicators, and sensitive content; and a consolidated analysis of KOPP and FA\_MED\_MMLU covering answer-key balance, extraction artefacts, translation limitations, and duplicates. It also adds 1,710 KOPP questions beyond the 168-question September 2023 session used in the conference papers, a unified evaluation with confidence intervals and significance tests, and an English--Persian comparison (Section~\ref{subsec6.3}). These additions shift the emphasis from the initial model demonstrations toward resource provenance, quality, limitations, and appropriate use. Some accuracy values differ from the conference papers because this article pools subject results by question count and rechecked answer extraction from raw model responses. In the conference papers, an LLM judge was asked to extract the option chosen in each response, and its extraction errors account for the discrepancies; the rechecked values reported here supersede the earlier values. The GPT-4o preference evaluation of free-form answers in~\cite{b1} is not repeated because we subsequently found GPT-4o unreliable as a judge of Persian medical text. The matched chain-of-thought self-consistency results from~\cite{b2} are included in Section~\ref{subsec6.2}, with their limitations stated explicitly.

The contributions are:
\begin{itemize}
  \item \textbf{A characterized Persian medical corpus.} A collection of 101,954 articles (85.5 million normalized words) from 11 Persian medical websites, analyzed for source imbalance, orthography, redundancy, boilerplate, provenance signals, and topical coverage.
  \item \textbf{MF3QA, a Persian patient--doctor QA resource.} A 24,000-pair dataset comprising 23,799 originally Persian pairs and 201 translated K-QA items, selected from a 186,029-pair raw crawl. We characterize source-specific splits, length-driven selection, QA alignment, repetition, overlap, privacy indicators, sensitive content, and topical differences.
  \item \textbf{A consolidated medical MCQ collection.} A collection of 1,878 KOPP questions from Iranian medical entrance examinations (including 1,710 questions added beyond the previously used September 2023 session) and 1,089 translated medical MMLU questions. We analyze answer-key balance, extraction artefacts, orthographic variation, translation limitations, and potential duplicates.
  \item \textbf{A translated training resource.} Approximately 18,000 Persian MedMCQA items produced by machine translation and filtered by two language-model evaluators. This resource is intended for training, not evaluation.
  \item \textbf{Two model-training case studies.} Documentation of Gaokerena-V, adapted through domain training and instruction tuning, and Gaokerena-R, trained with preference-based reasoning distillation. These experiments illustrate possible uses of the resource suite but do not establish that either recipe is uniformly superior to its backbone.
  \item \textbf{An English--Persian comparison.} Evaluation of Aya-Expanse-8B on corresponding English and Persian medical MMLU questions to measure the observed gap under the tested protocols and identify factors requiring further controlled study.
\end{itemize}

\section{Related Work}\label{sec2}

\subsection{English Small Medical Language Models}\label{subsec2.1}

Meerkat~\cite{b7} fine-tunes small models on medical chain-of-thought (CoT) reasoning paths extracted from textbooks, together with complementary biomedical data, emphasizing clinical reasoning rather than factual recall alone. MedS$^3$~\cite{b9} uses search and process supervision: Monte Carlo Tree Search over medical MCQs generates supervised fine-tuning, preference, and soft-label datasets, with a process reward model scoring intermediate steps. MedMobile~\cite{b10} targets deployment efficiency, fine-tuning Phi-3-mini~\cite{b11} on human-curated and synthetic medical data for use on resource-constrained devices.

\subsection{Persian Small Language Models}\label{subsec2.2}

Multilingual models support Persian, but Persian is a relatively small part of many training mixtures, and compact models may exhibit language mixing, grammatical errors, or limited linguistic nuance. Aya-Expanse~\cite{b12,b13} uses a multilingual training recipe that includes synthetic data from stronger models and offers Persian capability, but it is general-purpose rather than medically specialized. PersianMind~\cite{b14} is a bilingual Persian--English model developed by extending the LLaMA-2~\cite{b15} tokenizer and continuing training; it is also general-purpose. Sina-BERT~\cite{b6} addresses Persian medical text with an encoder model, but it is not a generative conversational model. Gaokerena builds on Aya-Expanse-8B and focuses on the availability and characterization of Persian medical resources, while using domain adaptation and preference-based training as two demonstrations of how such resources may be applied.

\section{Data Collection and Resource Characterization}\label{sec3}

The resource suite combines web-scraped Persian text, patient--doctor interactions, examination questions extracted from PDFs, and translated MCQs. These sources serve different purposes: the web corpus supplies domain text, MF3QA supplies conversational supervision, KOPP and FA\_MED\_MMLU support evaluation, and translated MedMCQA supplies training examples. The analyses below document both their scale and the limitations that affect how they should be used.

\subsection{Persian Medical Corpus}\label{subsec3.1}

\begin{figure}[ht]
\centering
\includegraphics[width=0.5\textwidth]{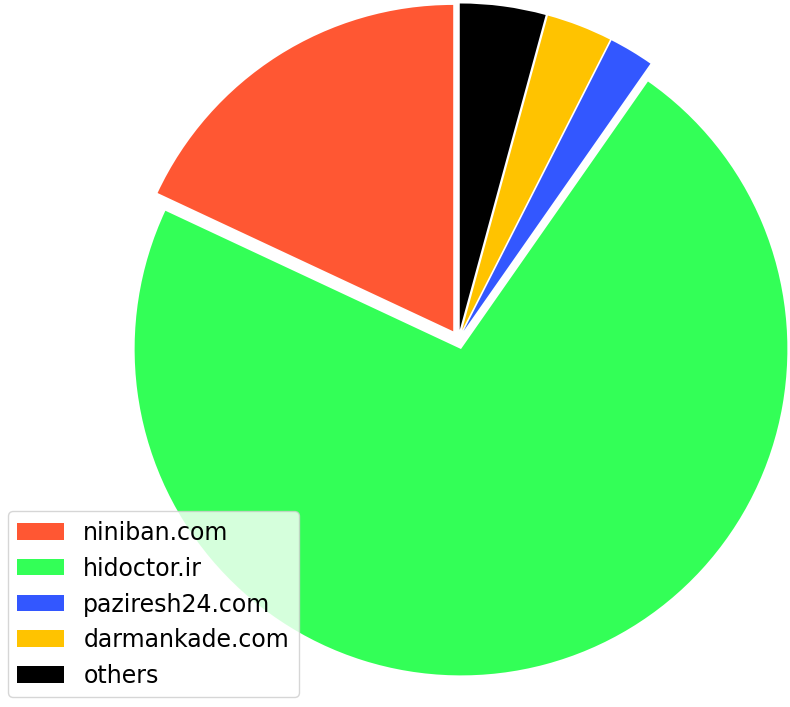}
\caption{Sources of the Persian medical corpus.}\label{fig1}
\Description{Bar chart showing the distribution of articles across 11 Persian medical websites, with hidoctor.ir dominating at 72.3 percent.}
\end{figure}

We compiled 101,954 articles totaling 85.5 million words from Persian medical magazines and health portals.\footnote{\url{https://huggingface.co/datasets/gaokerena/medical_corpus}} Our earlier publication~\cite{b1} reported approximately 90M tokens using the Hazm word tokenizer on raw text, which counts punctuation separately. The 85.5M figure reported here counts words after normalization and the word-boundary rule described below. The quantities are therefore not directly comparable. Garc\'ia-Ferrero et al.~\cite{b16} compiled medical text in English, Spanish, French, and Italian; Table~\ref{corpus_comparison} gives an indicative size comparison, noting that the published token counts and our word count use different counting conventions. Figure~\ref{fig1} shows the contribution of each source website.

\begin{table}[ht]
\caption{Comparison with corpora collected by Garc\'ia-Ferrero et al. Counts are reported in the source papers; our corpus is measured in words, so the comparison is indicative only.}\label{corpus_comparison}
\begin{tabular}{@{}lcl@{}}
\toprule
Language & Size & Collected by \\
\midrule
English & 1.1B & Garc\'ia-Ferrero et al. \\
Spanish & 950M & Garc\'ia-Ferrero et al. \\
French & 675M & Garc\'ia-Ferrero et al. \\
Italian & 143M & Garc\'ia-Ferrero et al. \\
Persian & 85.5M words & This work \\
\bottomrule
\end{tabular}
\end{table}

\subsubsection{Corpus Characteristics}\label{subsubsec:corpus_analysis}

We analyzed composition, orthographic quality, redundancy, boilerplate content, provenance, and topical coverage. Word counts use text normalized for Persian/Arabic character variants, digits, and punctuation, with the zero-width non-joiner (ZWNJ) treated as a word boundary.

\textbf{Composition.} The corpus contains 101,954 articles from 11 websites, 85.5M words, and 417M characters. The median article has 620 words (5th--95th percentile: 128--2,198). The distribution is highly skewed: hidoctor.ir supplies 72.3\% of articles and 62.0\% of words, and the two largest sources provide 80.7\% of words. Most smaller sources contain longer articles (median 1,013--2,246 words, compared with 527 for hidoctor.ir); niniban.com (737) and salammag.com (882) are exceptions. Only 502 articles (0.5\%) contain fewer than 50 words, and four exceed 10,000 words.

\begin{table}[ht]
\caption{Composition of the Persian medical corpus by source. Dup. is the share of articles flagged as exact or near duplicates; Boiler. is the mean share of article characters belonging to sentences that recur in at least 20 articles.}\label{tab:corpus_sources}
\small
\begin{tabular}{@{}lrrrrr@{}}
\toprule
Source & Articles & Words (M) & Median words & Dup. (\%) & Boiler. (\%) \\
\midrule
hidoctor.ir & 73,668 & 52.98 & 527 & 1.3 & 0.8 \\
niniban.com & 18,389 & 16.04 & 737 & 13.5 & 4.6 \\
darmankade.com & 3,316 & 5.90 & 1,731 & 13.1 & 2.9 \\
paziresh24.com & 2,259 & 3.43 & 1,307 & 0.0 & 9.2 \\
doctoreto.com & 2,071 & 4.38 & 2,246 & 0.2 & 1.4 \\
salammag.com & 787 & 0.77 & 882 & 0.0 & 4.3 \\
seebmagazine.com & 452 & 0.59 & 1,332 & 1.8 & 0.1 \\
majalepezeshki.ir & 322 & 0.41 & 1,183 & 0.0 & 0.0 \\
hamrah-teb.com & 312 & 0.45 & 1,506 & 0.0 & 0.4 \\
nabz.site & 270 & 0.42 & 1,555 & 0.0 & 0.0 \\
extern.ir & 108 & 0.13 & 1,013 & 0.9 & 0.0 \\
\midrule
Total & 101,954 & 85.50 & 620 & 3.8 & 1.8 \\
\bottomrule
\end{tabular}
\end{table}

\begin{figure}[ht]
\centering
\includegraphics[width=\textwidth]{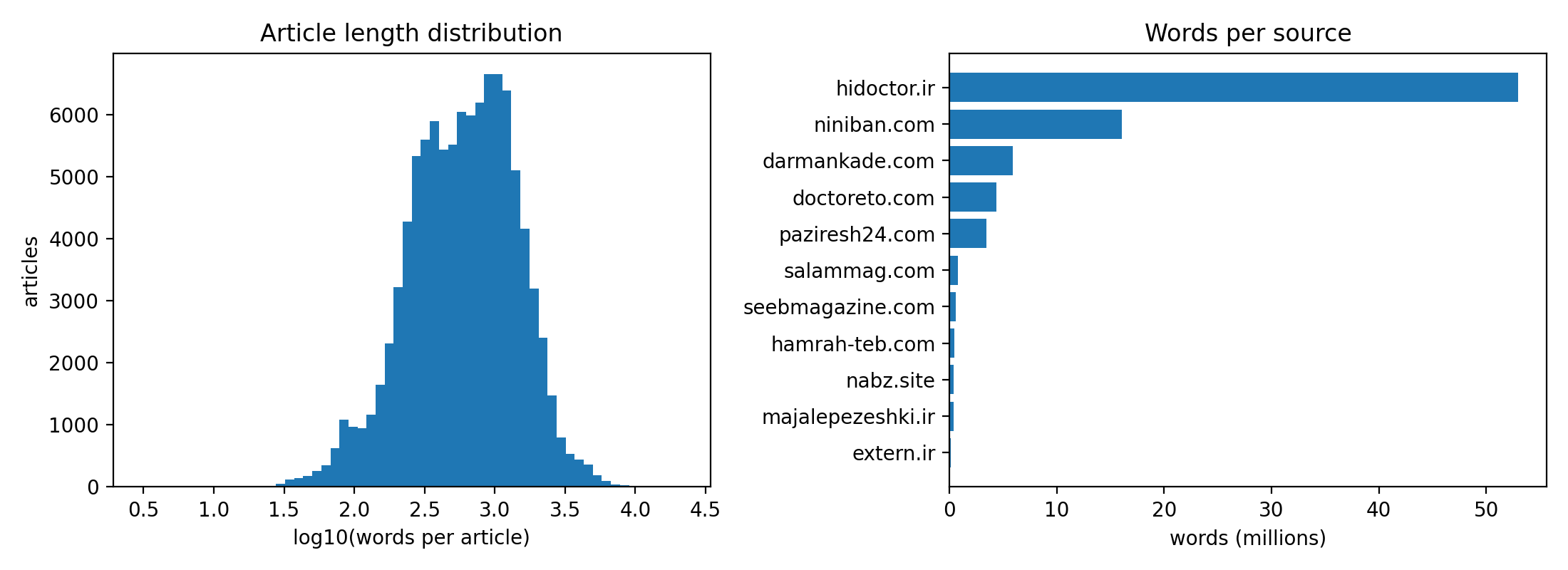}
\caption{Article length distribution (left) and words per source (right) in the Persian medical corpus.}\label{fig10}
\Description{Two-panel figure showing the distribution of article lengths and total word count per source.}
\end{figure}

\textbf{Script and orthography.} Persian/Arabic-script letters account for 96.6\% of non-whitespace characters and Latin letters for 0.6\%; 36 articles (0.04\%) are more than 20\% Latin. Orthographic variation remains substantial: Arabic Yeh or Kaf forms appear in 7.0\% of articles, 88\% of them from hidoctor.ir; 14.0\% of articles mix Persian and ASCII digits; and ZWNJ frequency ranges from 19 per 1,000 words at extern.ir to 96 at nabz.site. These differences fragment surface forms and support normalizing character variants before training.

\textbf{Redundancy.} After normalization, 2,257 articles (2.2\%) are exact duplicates of an earlier article and 3,903 (3.8\%, including exact duplicates) are exact or near duplicates. Near duplicates were detected using MinHash locality-sensitive hashing on five-word shingles of the first 500 words, with Jaccard threshold 0.8. Redundancy is concentrated in niniban.com (13.5\%) and darmankade.com (13.1\%), compared with 1.3\% for hidoctor.ir. Only 13 exact duplicates cross sources, but 423 articles have near duplicates in another source, consistent with syndication or light rewriting. Removing all flagged articles leaves 98,051 articles and 80.7M words, a 5.6\% reduction in words.

\textbf{Boilerplate and provenance.} We identified 2,193 distinct sentences appearing in at least 20 articles. They account for a mean of 1.8\% of article characters; 4,542 articles (4.5\%) exceed 10\%. Mean boilerplate shares are highest at paziresh24.com (9.2\%), niniban.com (4.6\%), and salammag.com (4.3\%). Frequent sentences include reuse and attribution footers, promotional prompts for booking or consultation, and standard drug-leaflet warnings. Attribution phrases also show that some content is translated or republished rather than original: a magazine translation-team credit occurs in 1,521 articles, a ``translated and rewritten by'' credit in 618, and a news-agency credit in 977; these counts may overlap.

\textbf{Topical coverage.} A 15-component TF-IDF/NMF topic model was fitted to a random sample of 20,000 articles. Each article was assigned to its dominant topic, and topic labels were manually assigned from top terms (Table~\ref{tab:corpus_topics}). The largest groups concern symptoms and conditions, cardiovascular disease and diabetes, and sleep and mental health. Some content is only loosely medical: cooking recipes (5.4\%) and directory listings for laboratories and clinics (2.7\%) together account for about 8\% of the sample, alongside cosmetics (6.5\%) and general lifestyle content. Topic distributions vary by source; niniban.com emphasizes pregnancy, childcare, and sexual health.

\begin{table}[ht]
\caption{Dominant topics in a 20,000-article sample of the corpus (NMF, 15 components).}\label{tab:corpus_topics}
\begin{tabular}{@{}lr@{}}
\toprule
Topic & Articles (\%) \\
\midrule
Symptoms, infections and surgery & 14.9 \\
Cardiovascular disease, diabetes and research news & 12.3 \\
Sleep, mental health and lifestyle & 12.0 \\
Diet and nutrition & 8.0 \\
Infant and child care, breastfeeding & 6.8 \\
Skin, hair and cosmetics & 6.5 \\
Medications & 5.9 \\
Sexual health and infertility & 5.9 \\
Pregnancy and fetal development & 5.7 \\
Cooking recipes & 5.4 \\
Cancer & 4.2 \\
Vitamins, minerals and supplements & 3.9 \\
Childbirth and delivery & 3.2 \\
Directory listings (labs, clinics, addresses) & 2.7 \\
Dental and oral health & 2.6 \\
\bottomrule
\end{tabular}
\end{table}

\subsection{MF3QA}\label{subsec3.2}

MF3QA (Medical Free-Form Farsi Question Answering) is a Persian patient--doctor QA dataset assembled from publicly accessible medical forums.\footnote{Cleaned dataset: \url{https://huggingface.co/datasets/gaokerena/MF3QA}. Raw crawl: \url{https://huggingface.co/datasets/gaokerena/MF3QA_uncleaned}.} Existing medical QA resources include ChatDoctor~\cite{b18}, CMtMedQA~\cite{b19}, DISC-Med-SFT~\cite{b20}, HuatuoGPT data~\cite{b21,b22}, MedDialog~\cite{b23}, and Medical-Meadow~\cite{b24}. MF3QA adds a Persian resource, although its selection procedure and source-specific limitations mean it should not be interpreted as a representative sample of all Persian medical interactions.

We collected 186,029 raw question--answer pairs and applied automated filtering and manual verification. The filtering is intentionally selective: approximately 87\% of the raw records were removed, primarily because many answers were brief, referral-only replies or otherwise too short to provide useful instruction-tuning targets. Before filtering, a portion of short answers (fewer than 50 tokens) was randomly dropped from the training and development pools to shift their length distribution; this omission was not applied to the test split. The test set also draws on Doctor-yab and Isovisit and includes 201 translated K-QA items~\cite{b25}. The final release contains 20,000 training, 2,000 development, and 2,000 test pairs. Personal-data screening was applied during cleaning, but the analysis below identifies residual indicators that should be reviewed before reuse or redistribution.

Drhast exposes a changing collection of roughly 2,000 recent records, each linking to related entries. We modeled the links as a graph and performed repeated breadth-first traversals, using newly posted records as roots and skipping previously visited records. This process collected 120,000 records from an estimated 200,000 over two weeks.

\begin{table}[ht]
\caption{Comparison with selected medical QA datasets. Record counts are approximate as reported by the respective sources.}\label{mf3qa_comparison}
\begin{tabular}{@{}llc@{}}
\toprule
Dataset & Language & Records \\
\midrule
ChatDoctor~\cite{b18} & English & 100K \\
CMtMedQA~\cite{b19} & Chinese & 68K \\
DISC-Med-SFT~\cite{b20} & Chinese & 465K \\
HuatuoGPT-sft-data-v1~\cite{b21} & Chinese & 226K \\
Huatuo-26M~\cite{b22} & Chinese & 26M \\
MedDialog~\cite{b23} & Chinese and English & 3.66M \\
Medical-Meadow~\cite{b24} & English & 160K \\
MF3QA (this work) & Persian & 24K \\
\bottomrule
\end{tabular}
\end{table}

\begin{figure}[ht]
\centering
\includegraphics[width=\textwidth]{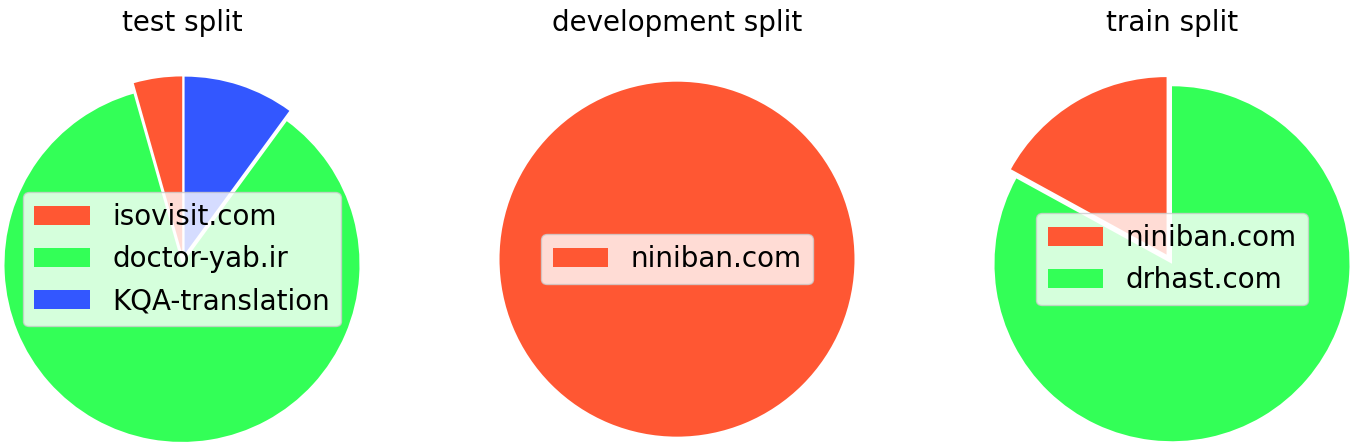}
\caption{Sources of the MF3QA question--answer pairs.}\label{fig2}
\Description{Pie chart showing the proportion of question-answer pairs collected from each Persian medical forum.}
\end{figure}

\subsubsection{MF3QA Characteristics}\label{subsubsec:mf3qa_analysis}

We compared the cleaned dataset with the raw crawl in terms of split composition, cleaning selection, length, question--answer alignment, answer style, privacy indicators, sensitive content, duplication, topical coverage, and overlap with the medical corpus. Lengths below are word counts on normalized text. Several style, privacy, sensitive-content, and topic screens rely on keywords or regular expressions and were not manually validated for precision; their counts are therefore approximate. Privacy and sensitive-content rates are calculated over questions unless stated otherwise.

\begin{figure}[ht]
\centering
\includegraphics[width=\textwidth]{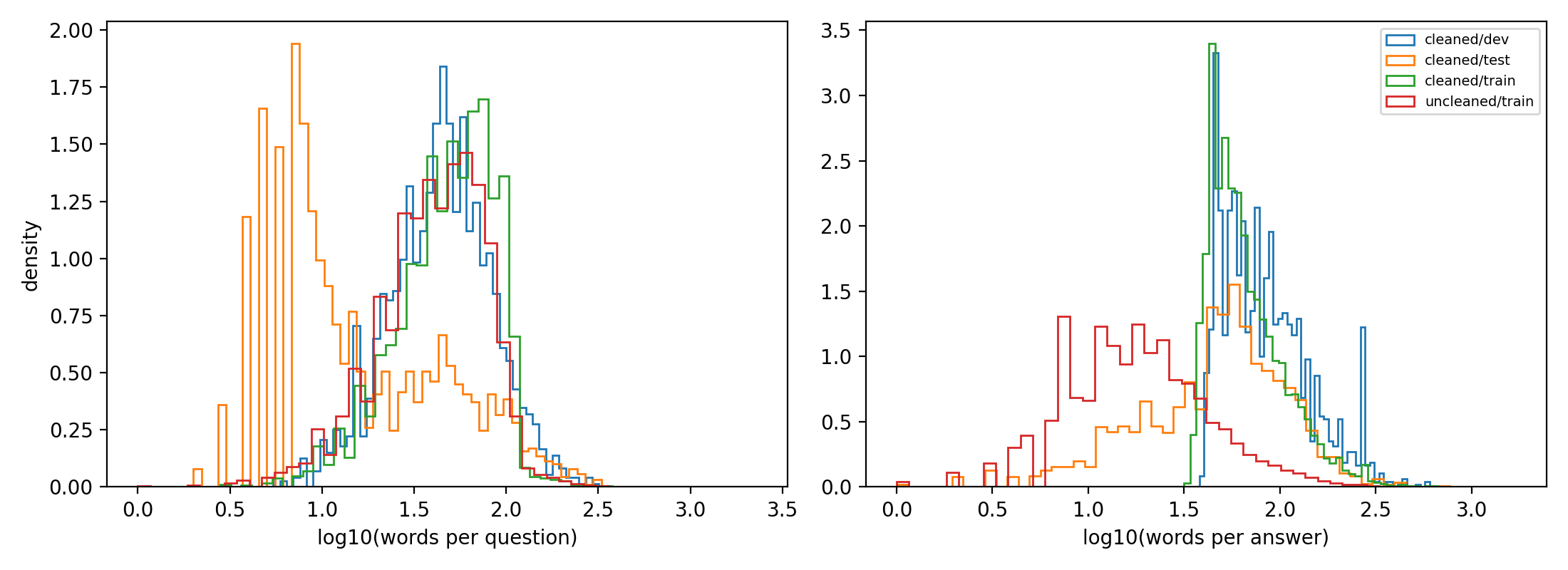}
\caption{Question length (left) and answer length (right), shown on a log scale, across MF3QA splits and the raw crawl.}\label{fig11}
\Description{Two-panel violin or box plots showing distributions of question and answer lengths across the MF3QA splits and raw crawl.}
\end{figure}

\textbf{Split composition.} The splits use different sources (Table~\ref{tab:mf3qa_splits}). Training combines Drhast (83.0\%) and Niniban (17.0\%); development comes entirely from Niniban; test comprises Doctor-yab (85.6\%), Isovisit (4.4\%), and translated K-QA items (10.1\%). Source is consequently confounded with split, and development does not represent the test distribution. Length also differs (Fig.~\ref{fig11}): median test question length is 11 words, compared with 51 in training and 45 in development. Two-sample Kolmogorov--Smirnov tests between training and test give $D=0.55$ for question length and $D=0.33$ for answer length ($p<10^{-170}$ for both). Results on MF3QA should therefore be reported by source and interpreted as out-of-distribution relative to the training set.

\begin{table}[ht]
\caption{MF3QA composition by split and source. Lengths are in words; Raw denotes the uncleaned crawl.}\label{tab:mf3qa_splits}
\small
\begin{tabular}{@{}llrrrr@{}}
\toprule
Split & Source & Rows & Q median & A median & A mean \\
\midrule
Train & drhast.com & 16,596 & 53 & 57 & 70.7 \\
Train & niniban.com & 3,404 & 46 & 76 & 97.5 \\
Dev & niniban.com & 2,000 & 45 & 77 & 97.8 \\
Test & doctor-yab.ir & 1,711 & 11 & 48 & 58.3 \\
Test & isovisit.com & 88 & 38 & 37 & 43.7 \\
Test & K-QA (translated) & 201 & 9 & 100 & 103.3 \\
\midrule
Raw & all sources & 186,029 & 43 & 17 & 25.9 \\
\bottomrule
\end{tabular}
\end{table}

\textbf{Selection effects of cleaning.} Cleaning retained 24,000 pairs from 186,029 raw records, with answer length dominating retention (Table~\ref{tab:mf3qa_funnel}). The median raw answer has 17 words; 74.3\% of raw answers contain fewer than 30 words, and 13.2\% are referral-only replies of at most 15 words. Of raw pairs with answers of at most 25 words (127,462, or 68.5\% of the crawl), 0.4\% were retained, compared with 80.6\% of answers longer than 50 words. Question length has a weaker relationship with retention (9.5--20.2\% across bins). Retention also varies by source: Drhast 13.6\%, Doctor-yab 3.1\%, and Niniban 58.9\%; Doctor-yab contributes only to test. Training and development therefore contain almost no short answers (0.015\% and 0.0\% below 30 words, respectively, and none below 50 tokens), whereas 27.9\% of test answers have fewer than 30 words. The random omission followed by filtering thus acts as an effective answer-length floor in training and development. Models tuned on these examples see longer responses than are typical of the source forums, and this mismatch should inform the interpretation of downstream outputs.

The raw crawl also contains scraping artefacts, including 83 pairs with empty questions or answers, repeated browser-error and media-player strings, and clinic advertisements. Of the 23,799 retained originally Persian pairs, 23,581 match a raw record by normalized question text and a further 170 by answer text, suggesting that some questions were edited during cleaning; 48 pairs could not be linked to the released crawl, three of which are Isovisit pairs added after the crawl was released. The 201 translated K-QA items are not in the crawl. The source-based retained counts total 23,799 Persian pairs, or 24,000 after adding K-QA. The answer-length bins in Table~\ref{tab:mf3qa_funnel} are computed independently on raw and retained answers rather than through record linkage, so the retention percentages are approximate. Word counts in this subsection use regular-expression word tokenization on normalized text.

\begin{table}[ht]
\caption{Raw MF3QA records and retained records by source and answer-length bin. Word counts use regular-expression word tokenization on normalized text. The 201 translated K-QA items were added separately and are not part of the raw crawl or of the length bins. Length bins are computed independently on raw and retained answers.}\label{tab:mf3qa_funnel}
\small
\begin{tabular}{@{}lrrr@{}}
\toprule
 & Raw & Kept & Kept (\%) \\
\midrule
\multicolumn{4}{@{}l}{\emph{By source}} \\
drhast.com & 122,233 & 16,596 & 13.58 \\
doctor-yab.ir & 54,539 & 1,711 & 3.14 \\
niniban.com & 9,172 & 5,404 & 58.92 \\
isovisit.com$^{\dagger}$ & 85 & 88 & --- \\
K-QA (translated) & --- & 201 & --- \\
\midrule
Total & 186,029 & 24,000 & \\
\midrule
\multicolumn{4}{@{}l}{\emph{By answer length (words), Persian pairs only}} \\
$\leq$10 & 50,508 & 137 & 0.3 \\
11--25 & 76,954 & 347 & 0.5 \\
26--50 & 38,776 & 7,371 & 19.0 \\
51--100 & 14,586 & 11,582 & 79.4 \\
101--200 & 4,385 & 3,623 & 82.6 \\
$>200$ & 820 & 739 & 90.1 \\
\midrule
Total & 186,029 & 23,799 & 12.8 \\
\bottomrule
\end{tabular}

\smallskip
\footnotesize $^{\dagger}$ Three Isovisit pairs in the cleaned dataset were added after the raw crawl was released, so the 85 Isovisit records in the crawl do not include them.
\end{table}

\textbf{Token counts.} Tokenization varies by model. With the Qwen2.5 tokenizer, training contains 7.28M tokens (3.18M in questions and 4.10M in answers, or 2.72 tokens per answer word); development contains 0.82M tokens and test 0.51M. Median answer lengths are 163, 204, and 135 tokens for training, development, and test, respectively. Some 14.95\% of test answers are shorter than 50 tokens, compared with none in training or development. Counts differ for other tokenizers, including the Aya-Expanse tokenizer used for training.

\begin{figure}[ht]
\centering
\includegraphics[width=0.7\textwidth]{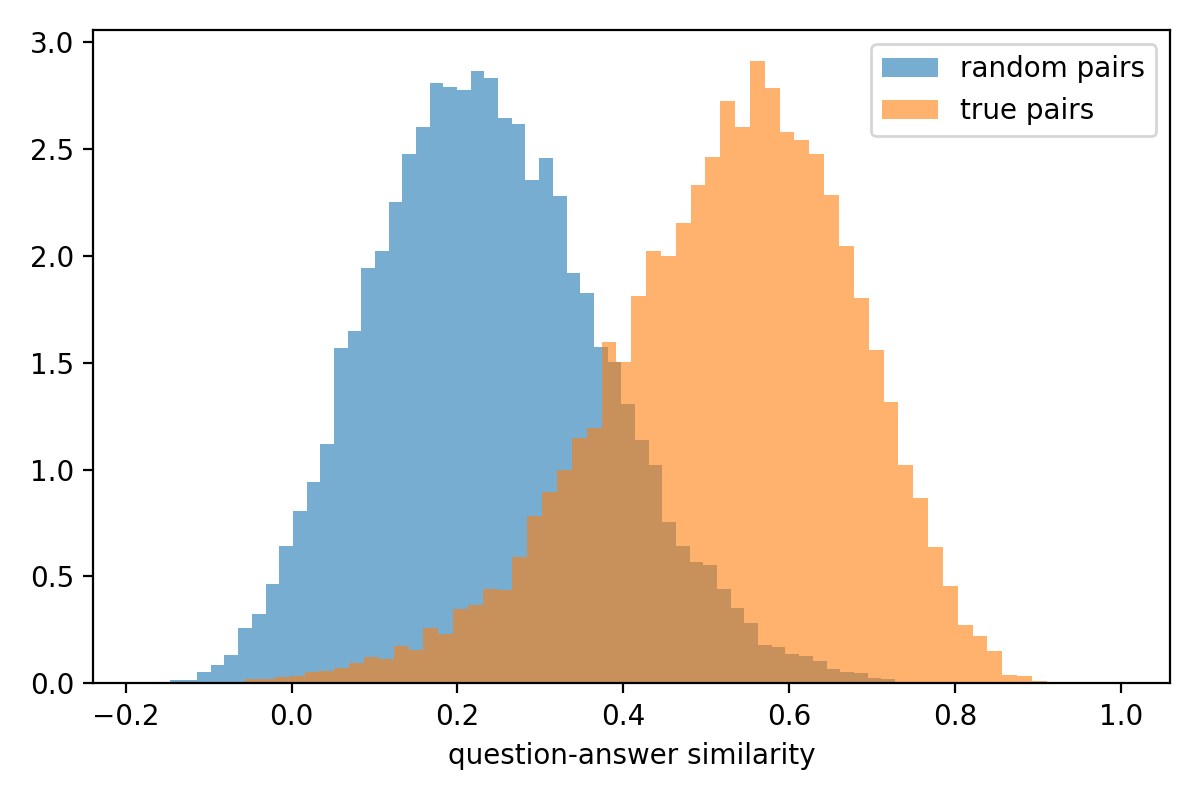}
\caption{Question--answer cosine similarity for true MF3QA pairs and randomly shuffled pairs.}\label{fig12}
\Description{Histogram comparing cosine similarity distributions for true question-answer pairs and randomly shuffled pairs.}
\end{figure}

\textbf{Question--answer alignment.} We scored each pair using cosine similarity from paraphrase-multilingual-MiniLM-L12-v2, truncating texts to 600 characters, and compared scores with randomly re-paired question--answer combinations (Fig.~\ref{fig12}). True pairs have higher mean similarity than random pairs (0.53 versus 0.24), although the distributions overlap. Using the median similarity of random pairs as a conservative threshold, 938 pairs (3.9\%) are flagged as potential mismatches: 3.9\% of training, 2.9\% of development, and 4.6\% of test. Rates by source are 2.5\% for Niniban, 4.2\% for Drhast, 4.6\% for Doctor-yab, 3.0\% for K-QA, and 8.0\% for Isovisit (88 pairs). A looser threshold, below the 95th percentile of random-pair similarity, flags 36.2\% overall and 41.3\% of Doctor-yab pairs. Because its questions are often short and answers generic, this is not an error rate. The embedding scores identify candidates for manual review, not confirmed mismatches.

\textbf{Style and provenance.} We used lexical proxies for answer style, including stock formal phrases, colloquial verb forms, greetings, and ZWNJ density. A combined ``LLM-like'' flag fires on 10.3\% of pairs; stock phrases alone occur in 7.9\%, and colloquial forms in 15.3\%. The combined flag is most frequent in translated K-QA (40.3\%, including 27.9\% with stock phrases and no colloquial forms), consistent with translation effects. Rates are 9.0\% for Drhast, 12.4\% for Niniban, 15.4\% for Doctor-yab, and 1.1\% for Isovisit. Registers also vary by source: only 0.3\% of Niniban answers contain colloquial forms and none starts with a greeting, whereas 20.2\% of Drhast answers are colloquial and 69.0\% start with a greeting. Formal medical language is common in human-authored text, so these proxies are not evidence of machine generation; they document source-dependent register differences.

\textbf{Duplication and leakage.} The cleaning pipeline did not deduplicate records. Within-split exact duplicates are uncommon: 17 training questions (0.09\%), one development question, and no test questions repeat an earlier question in the same split. Eleven training pairs and one development pair repeat both question and answer. Three development questions also occur in training (two with identical answers); no test question occurs in another split. MinHash LSH on three-word shingles for questions of at least five words (Jaccard threshold 0.8) finds about 77 near-duplicate question pairs within training, one within development, one within test, and 10 development questions with a near duplicate in training. No test question has a near duplicate in training. The nearest training question to a test question has median TF--IDF cosine similarity 0.16 (95th percentile 0.34); one test question (0.05\%) reaches 0.8, compared with 16 development questions (0.8\%). Answers are more repetitive: 3.7\% of cleaned pairs share an answer with at least four other pairs. The most frequent answer occurs 118 times and concerns fetal sex selection. Some repeated answers recommend home procedures without clear supporting evidence, illustrating that source authorship alone does not guarantee clinical soundness.

\textbf{Personal data.} The text is distributed largely as it appeared on public pages, so we screened for residual contact details using regular expressions for URLs, email addresses, mobile and landline numbers, and social-media handles. In raw answers, 1.08\% contain a mobile number, 0.58\% a landline number, 0.41\% a URL, and 0.08\% a social handle, often in clinic advertisements. In the cleaned dataset, at most 18 of 24,000 pairs contain one of these indicators: 12 through the question (including 11 in training and one email address in development) and six through the answer (four in training and two in test). A separate detector for a name following ``Dr.'' fires on 12.8\%, 9.2\%, and 5.7\% of training, development, and test questions, and on 2.8\%, 10.5\%, and 0.4\% of answers. The detector is approximate and may over-count forms of address, but physician signatures and names of treating physicians also occur. These checks do not establish full anonymization; further review and appropriate masking or release restrictions may be needed before redistribution.

\textbf{Sensitive content.} Keyword screens identify self-harm terms in 0.6\% of training questions (119) and 0.7\% of test questions (14); a first-person age below 18 in 0.8\% of training and 1.3\% of test; a child together with a dosing term in 5.6\% of training questions (1,111); and sexual-health terms in 4.2\% overall. The development split is atypical: 55.9\% of its questions mention pregnancy, versus 17.0\% in training and 2.8\% in test, because development comes entirely from Niniban. These heuristics may over- or under-count. Pediatric dosing and self-harm-related content require particular care; MF3QA-trained models should not be used for dosing advice, and flagged content should be reviewed before use in deployment-oriented settings.

\begin{table}[ht]
\caption{Keyword-screen flags on MF3QA questions. Percentages are per split, with counts in parentheses; screens are heuristic.}\label{tab:mf3qa_sensitive}
\small
\begin{tabular}{@{}lrrrr@{}}
\toprule
Flag & Train & Dev & Test & Raw crawl \\
\midrule
Rows & 20,000 & 2,000 & 2,000 & 186,029 \\
Self-harm term & 0.6 (119) & 0.0 (0) & 0.7 (14) & 0.5 (868) \\
First-person age $<18$ & 0.8 (154) & 0.5 (10) & 1.3 (25) & 1.1 (2,117) \\
Child + dosing term & 5.6 (1,111) & 7.9 (158) & 1.1 (22) & 3.9 (7,320) \\
Sexual-health term & 4.2 (840) & 6.7 (133) & 3.1 (62) & 6.9 (12,758) \\
Pregnancy term & 17.0 (3,405) & 55.9 (1,118) & 2.8 (56) & 15.7 (29,117) \\
\bottomrule
\end{tabular}
\end{table}

\textbf{Topical coverage.} A 12-component TF-IDF/NMF model was fitted to a random sample of 20,000 QA pairs using concatenated questions and answers; labels were assigned manually from top terms (Table~\ref{tab:mf3qa_topics}). The largest topic is better described as an informal first-person register concerning relationships, sexual concerns, and general complaints than as a single medical subject. Reproductive-health topics account for 22.6\% of the training sample, 69.8\% of development, and 11.0\% of test. General diagnosis, treatment, and surgery accounts for 40.0\% of test but 12.3\% of training and 1.9\% of development. This difference reinforces the need to avoid treating the development split as a proxy for test performance.

\begin{table}[ht]
\caption{Dominant topics in a 20,000-pair sample of MF3QA (NMF, 12 components), as a share of all sampled pairs and within each split (\%).}\label{tab:mf3qa_topics}
\small
\begin{tabular}{@{}lrrrr@{}}
\toprule
Topic & All & Train & Dev & Test \\
\midrule
Informal first-person narratives (relationships, general) & 23.5 & 26.1 & 5.1 & 14.9 \\
General diagnosis, treatment and surgery & 13.7 & 12.3 & 1.9 & 40.0 \\
Pain and physical complaints (dental, chest, back) & 13.2 & 14.1 & 3.3 & 14.8 \\
Medication use & 11.7 & 12.7 & 4.4 & 8.7 \\
Infertility and pregnancy planning & 7.5 & 6.5 & 22.2 & 2.4 \\
Weight, diet and growth & 6.7 & 6.6 & 4.6 & 9.7 \\
Pregnancy monitoring and miscarriage & 5.7 & 5.1 & 14.8 & 2.4 \\
Infant feeding and newborn care & 5.6 & 5.5 & 10.9 & 1.0 \\
Menstrual delay and pregnancy tests & 5.6 & 5.3 & 10.2 & 3.5 \\
Childbirth and caesarean & 3.0 & 2.6 & 8.7 & 1.8 \\
Ovulation and conception & 2.2 & 1.8 & 7.6 & 0.3 \\
Emergency contraception and menstrual disturbance & 1.7 & 1.3 & 6.4 & 0.6 \\
\bottomrule
\end{tabular}
\end{table}

\textbf{Overlap with the medical corpus.} Because Gaokerena-V was trained on the corpus and MF3QA, we measured normalized 8-gram overlap between QA text and all 101,954 corpus articles. This is an upper bound on overlap with training because only 60\% of the corpus was used for domain adaptation. At least one shared 8-gram occurs in 14.6\% of training, 41.0\% of development, and 18.5\% of test pairs. More than half of a pair's 8-grams overlap with corpus text for 2.2\%, 11.7\%, and 2.1\% of pairs, respectively. Shared boilerplate may explain some matches; individual matches were not manually verified. Development overlap is higher in part because its source, Niniban, also contributes to the corpus. The results in this article use MCQ benchmarks, but independent free-form evaluation on MF3QA would need to exclude or report separately pairs with substantial overlap.

\textbf{Implications for use.} MF3QA is most appropriately treated as a selected instruction-tuning resource whose training distribution differs from its test distribution. We recommend per-source test reporting, source-specific interpretation of the development split, review of privacy and sensitive-content flags, and care around highly repeated answers. The current analysis documents known limitations; it does not establish that every released record is clinically valid, representative, or fully anonymized.

\subsection{Persian Medical Multiple-Choice Questions}\label{subsec3.3}

Multiple-choice questions (MCQs) provide a structured evaluation format, although accuracy on MCQs is only a partial proxy for medical competence. We assembled two evaluation resources. KOPP contains 1,878 questions extracted from Iranian Basic Medical Science Entrance Exams from 2021--2024.\footnote{\url{https://huggingface.co/datasets/gaokerena/KOPP}} The September 2023 session contains 168 questions used in our earlier conference studies;IBMSEE and the KOPP refer to same benchmark ; the remaining 1,710 are added in this article. FA\_MED\_MMLU contains 1,089 manually translated questions from the medical portion of MMLU~\cite{b26}.\footnote{\url{https://huggingface.co/datasets/gaokerena/FA_MED_MMLU}} These two evaluation resources are released as a consolidated collection, but their source and construction differences remain relevant to analysis.

For training, we translated a portion of MedMCQA~\cite{b28} from English to Persian using DeepSeek-V3~\cite{b27}.\footnote{\url{https://github.com/Mehrdadghassabi/Gaokerena-R/tree/main/dataset/medmcqa_translation_filterd}} Items were sampled across disciplines including physiology, pharmacology, pathology, and clinical practice. Two language-model evaluators, Grok-3-mini~\cite{b29} and GPT-4.1-mini~\cite{b30}, independently rated semantic accuracy, fluency, and preservation of medical terminology. An item was retained only if both evaluators assigned a perfect score of five, yielding approximately 18,000 translated MCQs. The ratings were not validated against human annotators, so they should be interpreted as an automated filter rather than a human quality guarantee.

\subsubsection{MCQ Benchmark Characteristics}\label{subsubsec:mcq_analysis}

We analyzed size, subject coverage, answer-key balance, format, text quality, translation fidelity, and duplication. Quality screens based on regular expressions detect only specific artefacts and were not manually validated for precision.

\textbf{Size and composition.} KOPP contains 1,878 four-option questions, split into \texttt{sept23} (168 items) and \texttt{other\_years} (1,710 items). FA\_MED\_MMLU contains 1,089 items across six subjects; these counts match the corresponding English MMLU test subsets, so the Persian translation covers each selected subject's test partition. Together the resources contain 2,967 questions. All items have four options and a single designated answer, giving a nominal chance level of 25\%.

\begin{table}[ht]
\caption{Composition of the medical MCQ benchmarks.}\label{tab:mcq_composition}
\small
\begin{tabular}{@{}llrr@{}}
\toprule
Benchmark & Split / subject & Items & Share (\%) \\
\midrule
KOPP & sept23 & 168 & 8.9 \\
KOPP & other\_years & 1,710 & 91.1 \\
KOPP & all & 1,878 & 100.0 \\
\midrule
FA\_MED\_MMLU & anatomy & 135 & 12.4 \\
FA\_MED\_MMLU & clinical knowledge & 265 & 24.3 \\
FA\_MED\_MMLU & college biology & 144 & 13.2 \\
FA\_MED\_MMLU & college medicine & 173 & 15.9 \\
FA\_MED\_MMLU & medical genetics & 100 & 9.2 \\
FA\_MED\_MMLU & professional medicine & 272 & 25.0 \\
FA\_MED\_MMLU & all & 1,089 & 100.0 \\
\bottomrule
\end{tabular}
\end{table}

\textbf{Subject coverage.} The September 2023 KOPP session spans nine subjects: anatomy (32.1\%, 54 items), physiology (21.4\%), biochemistry and principles of health services (11.9\% each), bacteriology (9.5\%), parasitology (6.0\%), virology and mycology (3.0\% each), and entomology (1.2\%, two items). Some subjects are too small for separate inferential comparisons. The other sessions span 17 subjects after merging spelling variants, led by anatomy (27.0\%), physiology (17.9\%), bacteriology (10.8\%), and biochemistry (9.9\%). FA\_MED\_MMLU is dominated by professional medicine and clinical knowledge, which together comprise 49.3\% of its questions. Pooled scores should therefore be complemented by per-subject results.

\textbf{Answer key and format.} The correct option is in positions 1/2/3/4 for 32.1/26.8/16.1/25.0\% of September 2023 KOPP items, 27.3/24.6/25.0/23.2\% of other KOPP items, and 21.6/23.3/22.8/32.3\% of FA\_MED\_MMLU items. The departure from a uniform key is significant for the September 2023 session ($\chi^2$ test, $p=0.029$) and FA\_MED\_MMLU ($p=5.7\times10^{-7}$), but not for other KOPP sessions ($p=0.12$). A constant-position guess would score 32.1\%, 27.3\%, and 32.3\%, respectively. We did not determine whether the FA\_MED\_MMLU skew is inherited from the English source. The unique longest option is correct for 23.2\%, 20.8\%, and 25.3\% of items, respectively, when ties are excluded, so option length does not appear to provide a reliable cue. Median stem and option lengths are 13 and two words for KOPP and 17 and four words for FA\_MED\_MMLU. KOPP stores the answer as an integer index whereas FA\_MED\_MMLU uses a string label; harmonizing schemas would ease downstream use.

\textbf{Text quality.} PDF extraction introduces artefacts in KOPP, especially for right-to-left text containing Latin terms and numbers. Manual inspection found examples of reordered or truncated numerical values, stems copied from preceding options, appended stray text, unbalanced parentheses, missing whitespace at Persian--Latin boundaries, and mixed Arabic/Persian letter forms. These examples are qualitative, not prevalence estimates. Across 1,878 KOPP items, heuristic screens flag four items (0.2\%) with reordered or truncated numerals, two (0.1\%) whose stem equals an option from the same or previous question, 13 (0.7\%) with duplicate options, five (0.3\%) with unbalanced parentheses in stems, 124 (6.6\%) with no whitespace at a Persian--Latin boundary, and 605 (32.2\%) using Arabic Yeh or Kaf forms. The latter is 62.5\% in September 2023 versus 29.2\% in other sessions. Latin letters occur in 20.4\% of stems. Seventeen distinct subjects appear under 24 strings because of letter variants and invisible directional controls. The same screens flag 0.4\% of FA\_MED\_MMLU items for numeral artefacts, 0.2\% for duplicate options, 1.6\% for script-boundary issues, and 0.7\% for Arabic Yeh or Kaf. These screens are incomplete; flagged items should be reviewed before fine-grained comparison.

\textbf{Translation fidelity.} Some translated items leave anatomical and clinical terms in Latin script (e.g., \emph{levator veli palatini}), transliterate them, or translate terms inconsistently across options. Some ambiguity in the English wording may not have been preserved. We did not conduct a back-translation study or a second-annotator audit. A stratified human audit across the six subjects should measure how often translation changes the item's meaning or correct key. Annotation errors in the English MMLU subsets are also carried over to the translation.

\textbf{Duplication.} Matching normalized stems, 95 distinct stems occur more than once among the 2,967 questions, and 199 stem pairs have character $n$-gram TF--IDF cosine similarity of at least 0.8; six pairs cross between KOPP and FA\_MED\_MMLU. Since the comparison uses stems alone, short generic stems can match despite different options. These counts are therefore upper bounds on true duplicates, and the flagged identifiers should be manually reviewed.

\textbf{Implications for use.} KOPP is a session-structured benchmark, with only 168 questions in the September 2023 split, while FA\_MED\_MMLU is a complete translated test partition dominated by two subjects. We recommend per-subject and per-session reporting and, where appropriate, analyses both including and excluding repaired or flagged items.

\section{Gaokerena Models as Resource Applications}\label{sec4}

Both models use Aya-Expanse-8B~\cite{b12,b13} as their backbone. Gaokerena-V~\cite{b1} is obtained through domain adaptation on part of the medical corpus followed by instruction tuning on MF3QA. Gaokerena-R~\cite{b2} is post-trained on MCQs using preference-based reasoning distillation. The models illustrate different uses of the assembled resources, but the evaluation below does not isolate the effect of each resource or training stage.

\subsection{Gaokerena-V}\label{subsec4.1}

We adapted Aya-Expanse-8B in two LoRA~\cite{b31} stages. Domain adaptation used 60\% of the medical corpus, approximately 51M of the 85.5M normalized words, to expose the model to Persian medical terminology. The second stage used the MF3QA training split for instruction tuning and conversational QA. In total, the process used about 54 million new words; the earlier conference paper~\cite{b1} reported 57 million tokens under the previous tokenizer-based counting method. The MF3QA training split is source-selected and length-filtered, and part of the dataset overlaps with the full corpus as described in Section~\ref{subsubsec:mf3qa_analysis}.

\begin{table}[ht]
\caption{Gaokerena-V training hyperparameters.}\label{ft_it_training_details}
\small
\begin{tabular}{@{}lcc@{}}
\toprule
 & Domain adaptation & Instruction tuning \\
\midrule
Training data & medical\_corpus & MF3QA \\
Epochs & 1 & 1 \\
Batch size & 2 & 2 \\
Gradient accumulation steps & 16 & 16 \\
Optimizer & AdamW & AdamW \\
Learning rate & $5\times10^{-4}$ & $5\times10^{-4}$ \\
Maximum gradient norm & 0.3 & 0.3 \\
Warmup ratio & 0.03 & 0.03 \\
Weight decay & 0.1 & 0.5 \\
Maximum context length & 1024 & 1024 \\
Padding strategy & Left & Left \\
LoRA rank & 8 & 2 \\
LoRA alpha & 16 & 2 \\
LoRA dropout & 0.05 & 0.4 \\
Target modules & All linear layers & All linear layers \\
\bottomrule
\end{tabular}
\end{table}

\begin{figure}[ht]
\centering
\includegraphics[width=\textwidth]{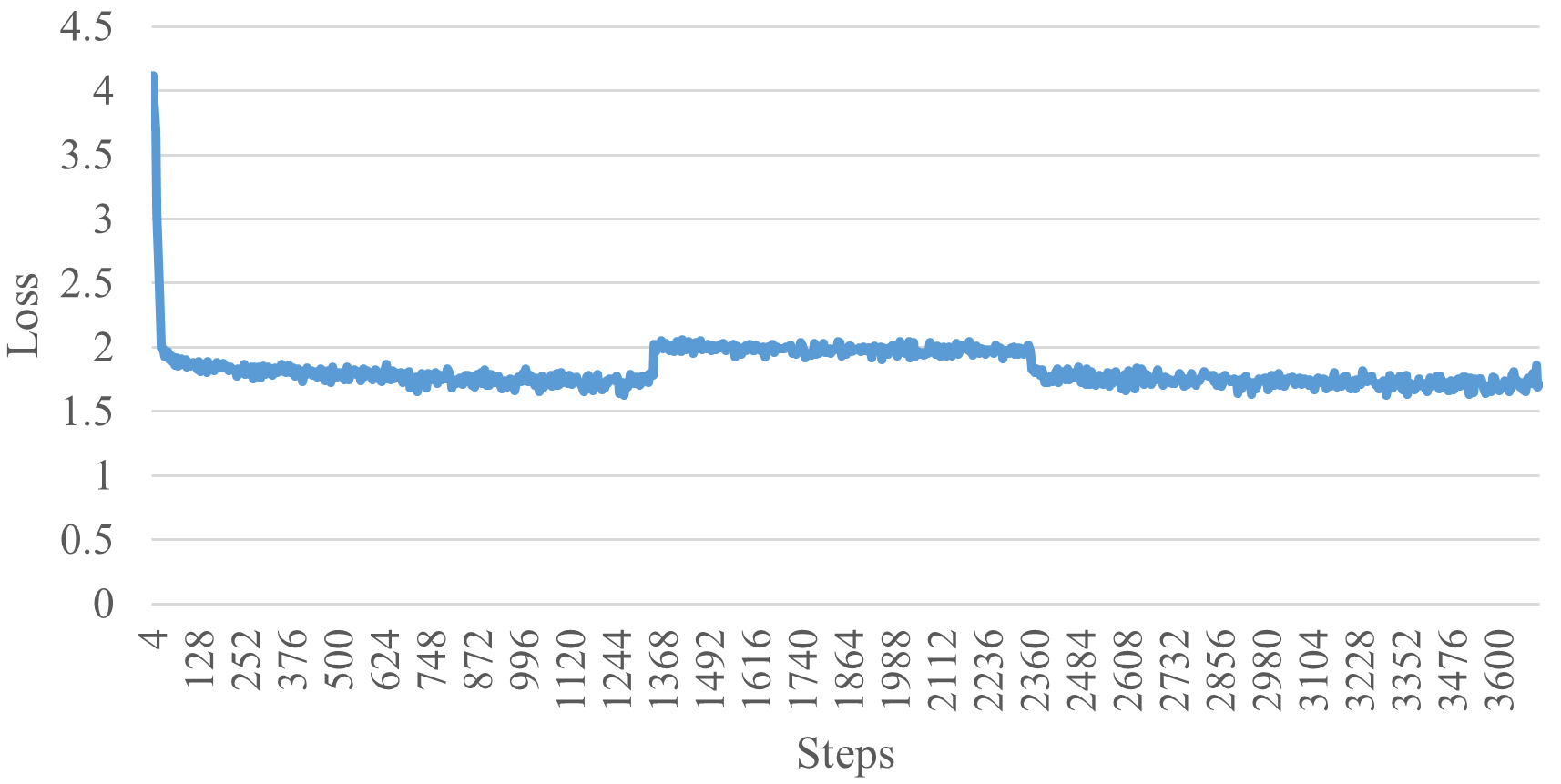}
\caption{Training loss during domain adaptation.}\label{fig3}
\Description{Line plot showing training loss decreasing over domain-adaptation steps.}
\end{figure}
\begin{figure}[ht]
\centering
\includegraphics[width=\textwidth]{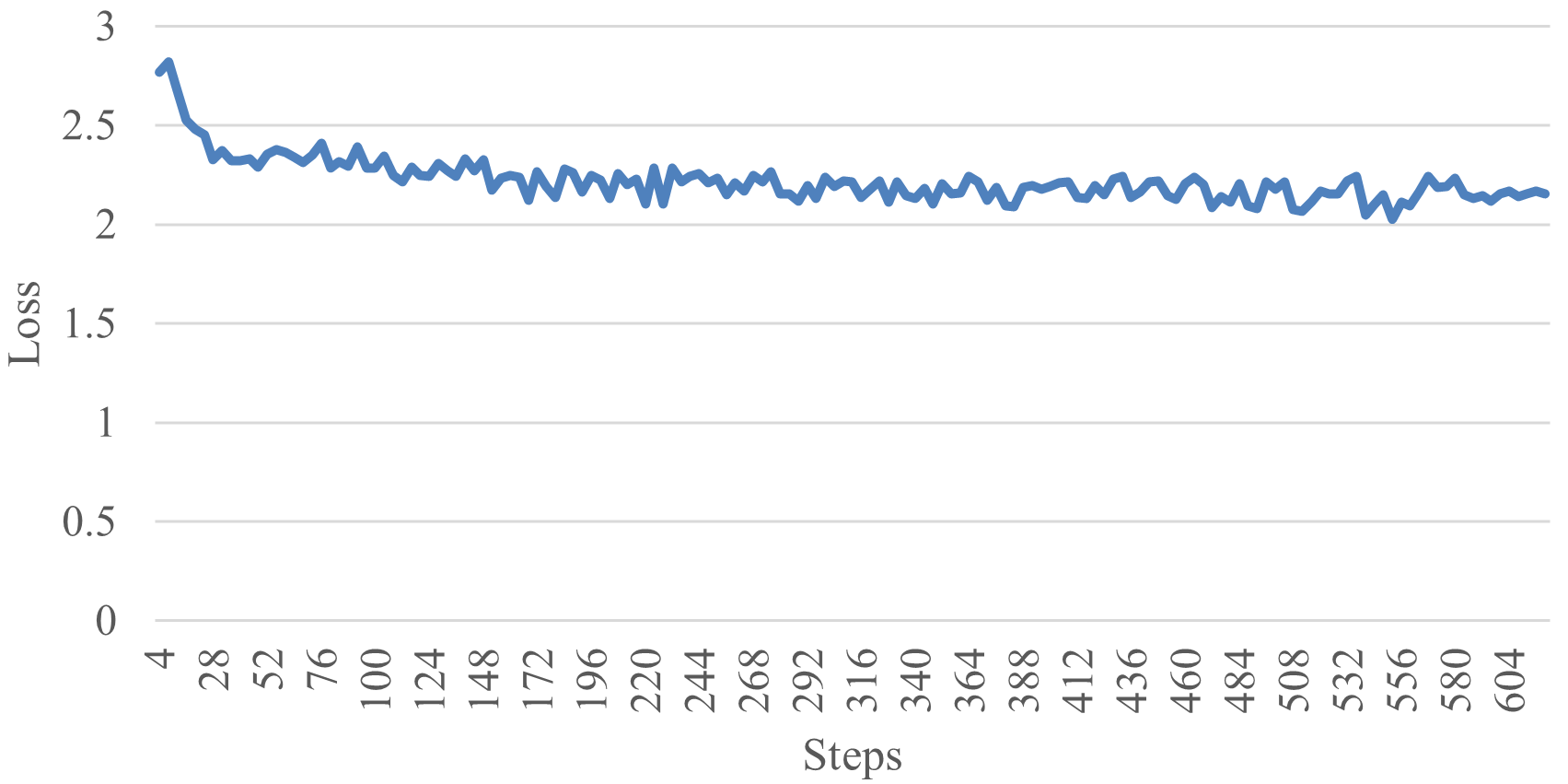}
\caption{Training loss during instruction tuning.}\label{fig4}
\Description{Line plot showing training loss decreasing over instruction-tuning steps.}
\end{figure}

\subsection{Gaokerena-R}\label{subsec4.2}

Gaokerena-R uses reinforcement learning from AI feedback (RLAIF)~\cite{b32} to create preference pairs $\mathcal{D}=\{(x,y_w,y_l)\}$, where $x$ is a medical question, $y_w$ is a preferred reasoning trajectory, and $y_l$ is a rejected response. The model is optimized using Direct Preference Optimization (DPO)~\cite{b33}. Aya-Expanse-8B is the student and DeepSeek-R1~\cite{b34} the teacher; questions are drawn from the translated MedMCQA training resource.

Two strategies construct preference pairs. In \textit{Expert-Guided Synthesis}, an incorrect initial student response $y_{\text{init}}$ becomes the rejected answer, while the teacher receives the ground-truth answer and writes a correct CoT explanation as the preferred answer. In \textit{Iterative Self-Correction}, the teacher critiques the initial response without revealing the answer; if the student's retry is correct, that retry becomes preferred and the initial answer rejected. The first strategy supplies approximately 19 pairs for every pair produced by the second. The final dataset contains about 11,000 pairs, with roughly 2 million preferred-response tokens and 2.5 million rejected-response tokens.

\begin{equation}
\mathcal{L}_{\text{DPO}}=-\mathbb{E}_{(x,y_w,y_l)\sim\mathcal{D}}\left[\log\sigma\left(\beta\log\frac{\pi_{\theta}(y_w\mid x)}{\pi_{\text{ref}}(y_w\mid x)}-\beta\log\frac{\pi_{\theta}(y_l\mid x)}{\pi_{\text{ref}}(y_l\mid x)}\right)\right]
\label{fr:DPOloss}
\end{equation}

Here $\sigma$ is the sigmoid function, $\pi_{\text{ref}}$ is the reference model, and $\beta$ controls the strength of the update relative to the reference policy. Minimizing the objective raises the probability of preferred responses relative to rejected responses. The available results evaluate the resulting model but do not include ablations that isolate the contribution of the two pair-generation strategies.

\begin{figure}[ht]
\centering
\includegraphics[width=\textwidth]{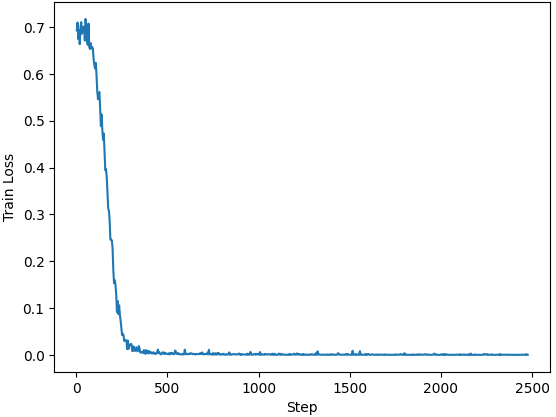}
\caption{DPO training loss.}\label{fig5}
\Description{Line plot showing DPO training loss over optimization steps.}
\end{figure}
\begin{figure}[ht]
\centering
\includegraphics[width=\textwidth]{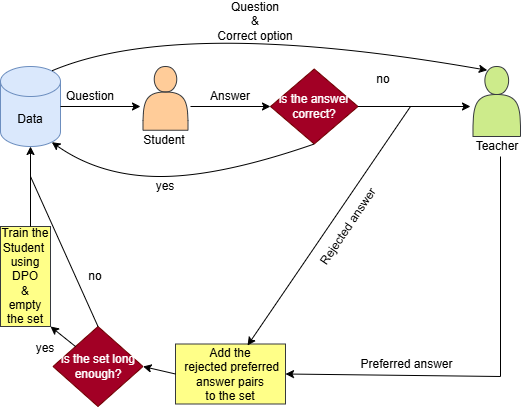}
\caption{Expert-Guided Synthesis preference-pair construction.}\label{fig6}
\Description{Flow diagram in which the student produces an incorrect response and the teacher, given the ground-truth answer, produces the preferred chain-of-thought response.}
\end{figure}
\begin{figure}[ht]
\centering
\includegraphics[width=\textwidth]{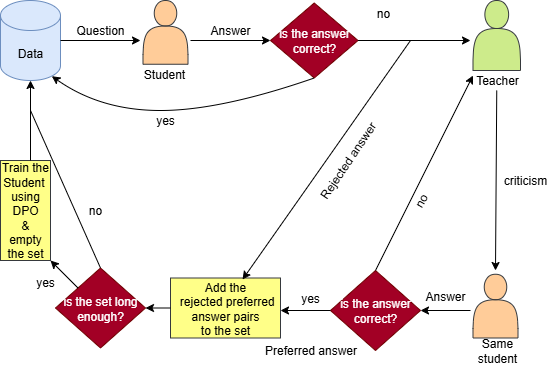}
\caption{Iterative Self-Correction preference-pair construction.}\label{fig7}
\Description{Flow diagram in which a teacher critiques a student response without revealing the answer, the student retries, and a correct retry becomes the preferred response.}
\end{figure}

\section{Companion Work: Claim-Level Uncertainty Heads}\label{sec:uhead}

A companion study~\cite{b3} investigates claim-level uncertainty heads for Gaokerena-V and Gaokerena-R. These auxiliary modules use internal model signals, including attention maps and token probabilities, to estimate whether a generated claim is supported or hallucinated, without external retrieval or repeated sampling. They do not change the backbone's generation process. The heads were trained using 1,600 Persian responses per backbone and obtained PR-AUC values of 0.4820 and 0.4652 (2.30 and 2.66 times the respective random baselines) and ROC-AUC values of 0.7852 and 0.7810 on held-out splits. In a preliminary analysis of five generations for each of the 168 September 2023 KOPP questions, all five generations selected the same option on only 14 Gaokerena-V questions and 37 Gaokerena-R questions. The full uncertainty-head methodology and evaluation are reported in~\cite{b3}; those results are separate from the resource characterization and do not affect the model comparisons below.

\section{Experiments and Results}\label{sec:results}

The experiments illustrate how the resources can be used to evaluate compact Persian medical models. They are not designed to estimate the independent causal contribution of each dataset or training stage. We evaluate on the six Persian-translated medical MMLU subsets (FA\_MED\_MMLU, 1,089 questions) and the September 2023 KOPP session (168 questions); other KOPP sessions are not evaluated in this article. Baselines are Aya-Expanse-8B, the backbone of both Gaokerena models; Qwen2.5~\cite{b35} (7.6B); and PersianMind~\cite{b14}. Results come from one evaluation run per configuration. Except for Table~\ref{tab:cotsc}, accuracies are calculated from per-question predictions; Table~\ref{tab:cotsc} is pooled by question count from subject-level results reported in~\cite{b2}.

\subsection{Evaluation Setup and Statistical Methods}\label{subsec6.0}

We report accuracy and 95\% Wilson score intervals. The original pairwise comparisons use two-sided unpaired two-proportion $z$-tests; because systems answer shared questions, we additionally conducted exact two-sided McNemar tests on paired per-question correctness indicators for (i) direct-prompting Gaokerena-V versus Aya-Expanse-8B and (ii) Gaokerena-R versus Aya-Expanse-8B under matched five-sample CoT self-consistency without a verifier. Pairing was based on subject, question text, and gold answer. Let $b$ denote items answered correctly only by Gaokerena-V and $c$ items answered correctly only by Aya-Expanse-8B; the exact McNemar test uses the binomial distribution conditional on $b+c$. The paired results are summarized in Table~\ref{tab:mcnemar}. The comparisons reported in Tables~\ref{tab:significance} and~\ref{tab:cotsc_sig} were selected after results were observed, not pre-registered. The tables show uncorrected $p$-values; Holm corrections are reported for the ten comparisons in Table~\ref{tab:significance} and the six comparisons in Table~\ref{tab:cotsc_sig}. Accuracy against the 36\% entrance-exam pass mark is tested with a one-sided exact binomial test. Only pooled differences are tested; per-subject comparisons are descriptive. These tests capture uncertainty across questions, not across model-training or decoding seeds. English--Persian comparisons use the same unpaired approach, and some Persian counts are reconstructed from rounded percentages.

\subsection{Gaokerena-V Results}\label{subsec6.1}

Table~\ref{tab:model_results_vs_general_purpose_languages} reports direct, zero-shot prompting results. Gaokerena-V has the highest pooled MMLU accuracy among the directly prompted systems (49.31\%, 537/1,089), compared with 46.65\% (508/1,089) for Aya-Expanse-8B and 44.81\% (488/1,089) for Qwen2.5. It has the highest score in five of the six subjects; Gaokerena-R is highest on college medicine. However, the difference between Gaokerena-V and its backbone is 2.66 percentage points and is not statistically significant under the reported unpaired test ($p=0.21$). As a paired sensitivity analysis, Gaokerena-V was correct on 537 items and Aya-Expanse-8B on 508; there were 117 items correct only for Gaokerena-V and 88 correct only for Aya-Expanse-8B. The exact two-sided McNemar test yielded $p=0.0502$, just above the conventional 0.05 threshold. This near-threshold result remains statistically inconclusive and should be interpreted cautiously because the comparisons are exploratory. Its 4.50-point difference from Qwen2.5 has uncorrected $p=0.035$ but does not survive the stated Holm correction. These results are compatible with a modest improvement, but do not establish that domain adaptation and instruction tuning caused one.

On the September 2023 KOPP session, Gaokerena-V scores 38.69\% (65/168), above the nominal 36\% pass mark. Its 95\% Wilson interval is 31.66--46.23\%, which includes the pass mark, and the one-sided exact binomial test is not significant ($p=0.26$). It answers seven more questions correctly than Aya-Expanse-8B, a difference of 4.17 points ($p=0.43$ under the unpaired test). Because a constant-position guess reaches 32.1\% on this session due to answer-key imbalance (Section~\ref{subsubsec:mcq_analysis}), the result should be interpreted as a point estimate on one small session, not as evidence of exam readiness.

The largest subject-level gain over the backbone is in anatomy (7.4 points, or 10 additional correct answers out of 135); this comparison is descriptive and was not tested. A possible explanation is that the corpus contains relevant anatomy material, but this was not tested. PersianMind is fastest in the reported setup, but its pooled accuracy is 25.90\% (95\% interval 23.38--28.58\%) and its tendency to produce very short answers may affect answer extraction.

\begin{table}[ht]
\caption{Direct (zero-shot) prompting results. Accuracy is in percent; the MMLU score is pooled over 1,089 questions. Confidence intervals and tests are in Tables~\ref{tab:ci} and~\ref{tab:significance}.}\label{tab:model_results_vs_general_purpose_languages}
\scriptsize
\setlength{\tabcolsep}{3pt}
\begin{tabular}{@{}lccccc@{}}
\toprule
 & Gaokerena-V & Gaokerena-R & Aya-Expanse-8B & Qwen2.5 & PersianMind \\
 & (ours) & (ours) & (backbone) & & \\
\midrule
Anatomy & \textbf{48.15} & 41.48 & 40.74 & 41.48 & 25.19 \\
Medical genetics & \textbf{53.00} & 49.00 & 49.00 & 52.00 & 34.00 \\
College medicine & 43.93 & \textbf{46.24} & 44.51 & 43.35 & 20.23 \\
Clinical knowledge & \textbf{55.47} & 52.45 & 52.08 & 47.92 & 25.28 \\
Professional medicine & \textbf{47.06} & 41.91 & 45.59 & 43.01 & 23.90 \\
College biology & \textbf{47.22} & 44.44 & 45.14 & 42.36 & 32.64 \\
MMLU avg. (pooled) & \textbf{49.31} & 46.10 & 46.65 & 44.81 & 25.90 \\
IBMSEE Sept. 2023 & \textbf{38.69} & 35.12 & 34.52 & 33.33 & 19.64 \\
\midrule
Parameters & 8B & 8B & 8B & 7.6B & 6.8B \\
Prompt type & Direct & Direct & Direct & Direct & Direct \\
Inference time & $\approx$10 s & $\approx$10 s & $\approx$10 s & $\approx$15 s & $\approx$2 s \\
\bottomrule
\end{tabular}
\end{table}

\begin{table}[ht]
\caption{Pooled accuracy with 95\% Wilson confidence intervals. MMLU $n=1,089$; KOPP September 2023 $n=168$. The Gaokerena-R CoT-SC-plus-verifier configuration is not compute-matched to direct baselines.}\label{tab:ci}
\small
\begin{tabular}{@{}lcc@{}}
\toprule
System & MMLU pooled (95\% CI) & KOPP Sept. 2023 (95\% CI) \\
\midrule
Gaokerena-V (direct) & 49.31 [46.35, 52.28] & 38.69 [31.66, 46.23] \\
Gaokerena-R (direct) & 46.10 [43.16, 49.07] & 35.12 [28.31, 42.60] \\
Gaokerena-R (CoT-SC + verifier) & 52.80 [49.83, 55.75] & 46.43 [39.05, 53.97] \\
Aya-Expanse-8B (direct) & 46.65 [43.70, 49.62] & 34.52 [27.75, 41.99] \\
Qwen2.5 (direct) & 44.81 [41.88, 47.78] & 33.33 [26.65, 40.76] \\
PersianMind (direct) & 25.90 [23.38, 28.58] & 19.64 [14.34, 26.30] \\
\bottomrule
\end{tabular}
\end{table}

\begin{table}[ht]
\caption{Pairwise accuracy differences (percentage points) and uncorrected $p$-values from unpaired two-proportion $z$-tests. The reported Holm correction covers ten tests; only the MMLU comparison of Gaokerena-R CoT-SC plus verifier against the directly prompted backbone remains significant. Reanalysis with paired per-question tests is recommended.}\label{tab:significance}
\small
\begin{tabular}{@{}lrrrr@{}}
\toprule
 & \multicolumn{2}{c}{MMLU pooled} & \multicolumn{2}{c}{KOPP Sept. 2023} \\
\cmidrule(lr){2-3}\cmidrule(lr){4-5}
Comparison & $\Delta$ & $p$ & $\Delta$ & $p$ \\
\midrule
Gaokerena-V vs. Aya-Expanse-8B (direct) & +2.66 & 0.21 & +4.17 & 0.43 \\
Gaokerena-V vs. Qwen2.5 (direct) & +4.50 & 0.035 & +5.36 & 0.31 \\
Gaokerena-R (CoT-SC + verifier) vs. Gaokerena-V (direct) & +3.49 & 0.10 & +7.74 & 0.15 \\
Gaokerena-R (CoT-SC + verifier) vs. Aya-Expanse-8B (direct) & +6.15 & 0.004$^{*}$ & +11.90 & 0.026 \\
Gaokerena-R (direct) vs. Aya-Expanse-8B (direct) & -0.55 & 0.80 & +0.60 & 0.91 \\
\bottomrule
\end{tabular}
\end{table}

\subsection{Gaokerena-R Results}\label{subsec6.2}

Gaokerena-R is designed for reasoning-style output, so we evaluate it with chain-of-thought (CoT) prompting as well as direct prompting. Under direct prompting, it does not outperform its backbone (46.10\% versus 46.65\%, reported $p=0.80$).

\paragraph{Matched CoT self-consistency.} Each model receives a CoT prompt and generates five answers per question at temperature 1.0. A self-consistency rule~\cite{b36} accepts an option only if at least three responses agree; otherwise the item is unanswered and counted as incorrect. No verifier is used. Gaokerena-R obtains 48.58\% pooled MMLU accuracy (529/1,089), compared with 46.92\% for Aya-Expanse-8B (511/1,089; +1.65 points). The unpaired two-proportion test gives $p=0.44$. On the same 1,089 paired questions, Gaokerena-R is correct when Aya-Expanse-8B is incorrect on 93 items, while the reverse occurs on 75 items; the exact two-sided McNemar test is not significant ($p=0.190$). Gaokerena-V obtains 40.13\%. The reported difference between Gaokerena-R and Gaokerena-V is significant ($+8.45$ points, $p<0.001$), while the Gaokerena-R/backbone difference is not significant on MMLU or KOPP. Gaokerena-V scores 6.80 points below its backbone under this protocol (reported $p=0.001$). This may reflect greater abstention under the three-vote rule, but coverage was not recorded separately. Under matched inference computation, the results do not establish a DPO improvement over the backbone, though Gaokerena-R has a higher point estimate than Gaokerena-V in this setting.

\begin{table}[ht]
\caption{Matched CoT self-consistency comparison: five samples at temperature 1.0 and a three-vote majority, without a verifier. Items without a majority are counted as incorrect. Values were pooled from subject-level results in~\cite{b2}; one Gaokerena-R subject value (Clinical knowledge) was corrected after rechecking answer extraction, as described in Section~\ref{sec1}.}\label{tab:cotsc}
\small
\begin{tabular}{@{}lccc@{}}
\toprule
 & Gaokerena-R & Gaokerena-V & Aya-Expanse-8B \\
\midrule
Anatomy & \textbf{42.22} & 39.26 & 40.74 \\
Medical genetics & \textbf{50.00} & 41.00 & 45.00 \\
College medicine & 47.98 & 37.57 & \textbf{48.55} \\
Clinical knowledge & \textbf{55.47} & 46.79 & 54.72 \\
Professional medicine & \textbf{44.85} & 37.13 & 43.75 \\
College biology & \textbf{48.61} & 36.81 & 43.75 \\
MMLU avg. (pooled) & \textbf{48.58} & 40.13 & 46.92 \\
\quad 95\% CI & [45.62, 51.54] & [37.26, 43.07] & [43.98, 49.89] \\
IBMSEE Sept. 2023 & \textbf{38.69} & 29.76 & 35.71 \\
\quad 95\% CI & [31.66, 46.23] & [23.36, 37.07] & [28.86, 43.21] \\
\midrule
Prompt & CoT & CoT & CoT \\
Decoding & CoT-SC, no verifier & CoT-SC, no verifier & CoT-SC, no verifier \\
Generation time & $\approx5\times35$ s & $\approx5\times35$ s & $\approx5\times35$ s \\
\bottomrule
\end{tabular}
\end{table}

\begin{table}[ht]
\caption{Pairwise comparisons among matched CoT-SC systems. Differences are percentage points; $p$-values are uncorrected unpaired two-proportion $z$-tests.}\label{tab:cotsc_sig}
\small
\begin{tabular}{@{}lrrrr@{}}
\toprule
 & \multicolumn{2}{c}{MMLU pooled} & \multicolumn{2}{c}{KOPP Sept. 2023} \\
\cmidrule(lr){2-3}\cmidrule(lr){4-5}
Comparison & $\Delta$ & $p$ & $\Delta$ & $p$ \\
\midrule
Gaokerena-R vs. Aya-Expanse-8B & +1.65 & 0.44 & +2.98 & 0.57 \\
Gaokerena-R vs. Gaokerena-V & +8.45 & $<0.001$ & +8.93 & 0.085 \\
Aya-Expanse-8B vs. Gaokerena-V & +6.80 & 0.001 & +5.95 & 0.25 \\
\bottomrule
\end{tabular}
\end{table}

\begin{table}[ht]
\caption{Paired exact McNemar tests comparing each Gaokerena model with Aya-Expanse-8B on shared medical MMLU questions. $b$ and $c$ are discordant pairs: questions answered correctly only by the first-listed model and only by Aya-Expanse-8B, respectively. $p$-values are two-sided and unadjusted.}
\label{tab:mcnemar}
\small
\begin{tabular}{@{}lrrr@{}}
\toprule
Comparison & $b$ & $c$ & Exact $p$ \\
\midrule
Gaokerena-V (direct) vs. Aya-Expanse-8B (direct) & 117 & 88 & 0.0502 \\
Gaokerena-R (CoT-SC, no verifier) vs. Aya-Expanse-8B (CoT-SC, no verifier) & 93 & 75 & 0.190 \\
\bottomrule
\end{tabular}
\end{table}

\paragraph{Adding a verifier.} In the preferred Gaokerena-R configuration, questions without a three-vote majority are passed to Aya-Expanse-8B, the backbone, as a verifier. Accuracy rises from 48.58\% to 52.80\% on pooled MMLU (575/1,089, 46 additional correct answers) and reaches 46.43\% on KOPP. Reported inference time is approximately 193 seconds per question, about 19 times the direct baselines (about 10 seconds). Against directly prompted Aya-Expanse-8B, the reported MMLU difference is +6.15 points ($p=0.004$, surviving the stated Holm correction); the KOPP difference is +11.90 points ($p=0.026$ before correction). The difference from directly prompted Gaokerena-V is not significant in either benchmark ($p=0.10$ and $p=0.15$). The reported KOPP interval is 39.05--53.97\%, and the one-sided exact test against the 36\% pass mark gives $p=0.003$.

This comparison is not compute-matched. Five samples, CoT prompting, and a verifier provide more inference computation than direct prompting, and the backbone was not run through the identical CoT-SC-plus-verifier pipeline. The 52.80\% result therefore describes the best tested system configuration, not the isolated contribution of DPO or of any specific training resource.

\paragraph{Why direct and CoT results differ.} Pooled MMLU accuracy rises from 46.10\% to 48.58\% for Gaokerena-R (+2.5 points), changes from 46.65\% to 46.92\% for the backbone (+0.3), and falls from 49.31\% to 40.13\% for Gaokerena-V (-9.2). These settings differ in sample count, temperature, and treatment of unanswered questions, so the results do not identify the cause. Plausible explanations include a closer match between Gaokerena-R's reasoning-style training targets and CoT prompting, greater sample variability for Gaokerena-V, error propagation during longer reasoning outputs, and failures to produce a parsable final option. We did not measure coverage or extraction failure rates, so these explanations remain hypotheses. Future comparisons should report per-sample accuracy, abstention/coverage, answer-extraction failures, and direct versus CoT results under better-matched decoding settings.

\begin{table}[ht]
\caption{Gaokerena models under their preferred tested prompting strategies. The configurations are not compute-matched; Table~\ref{tab:cotsc} is the matched comparison without a verifier. Accuracy is in percent.}\label{tab:med_opns_comparison}
\small
\begin{tabular*}{\textwidth}{@{\extracolsep\fill}lccc}
\toprule
 & Gaokerena-R & Gaokerena-V & Aya-Expanse-8B \\
\midrule
Anatomy & 47.41 & \textbf{48.15} & 40.74 \\
Medical genetics & \textbf{56.00} & 53.00 & 49.00 \\
College medicine & \textbf{50.29} & 43.93 & 44.51 \\
Clinical knowledge & \textbf{58.87} & 55.47 & 52.08 \\
Professional medicine & \textbf{48.90} & 47.06 & 45.59 \\
College biology & \textbf{54.86} & 47.22 & 45.14 \\
MMLU avg. (pooled) & \textbf{52.80} & 49.31 & 46.65 \\
IBMSEE Sept. 2023 & \textbf{46.43} & 38.69 & 34.52 \\
\midrule
Prompt & CoT & Direct & Direct \\
Decoding & CoT-SC + verifier & Zero-shot & Zero-shot \\
Inference time & $\approx5\times35+10+8$ s & $\approx10$ s & $\approx10$ s \\
\bottomrule
\end{tabular*}
\end{table}

\begin{figure}[ht]
\centering
\includegraphics[width=\textwidth]{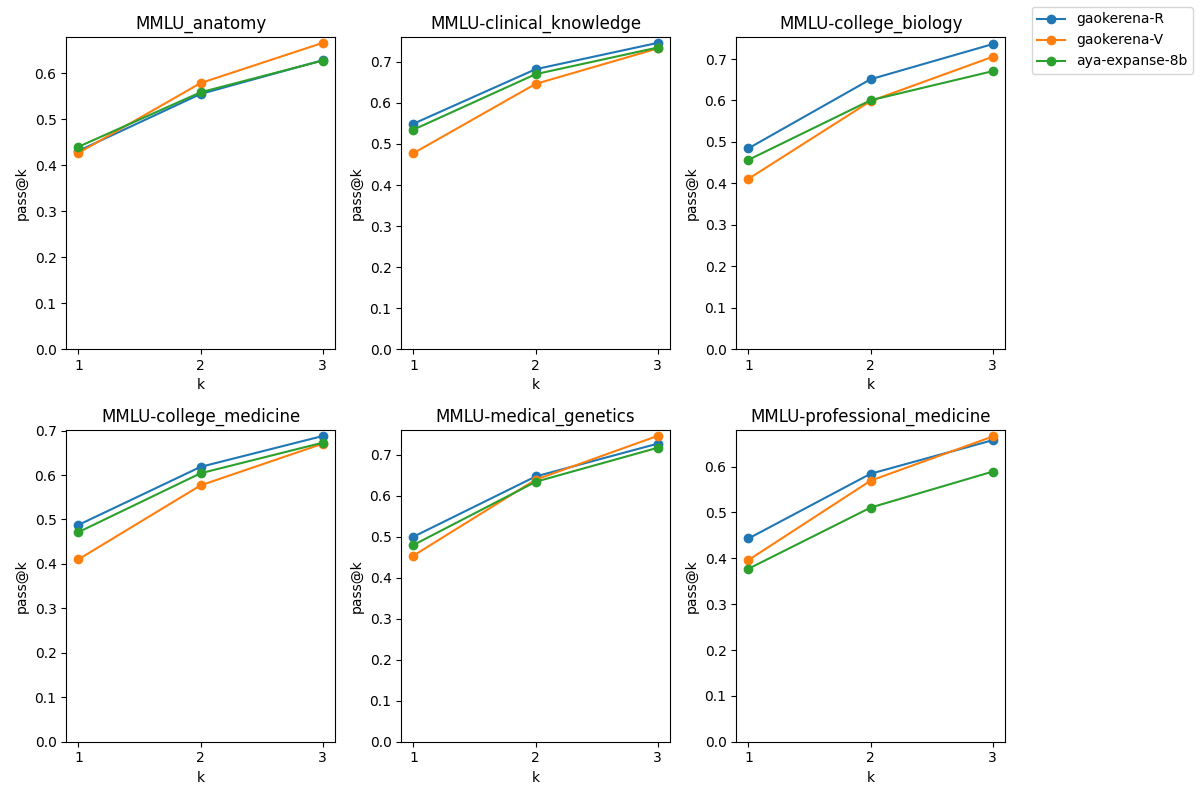}
\caption{Pass@k on the Persian-translated medical MMLU subsets.}\label{fig8}
\Description{Line plot showing pass@k scores for Gaokerena-R, Gaokerena-V, and Aya-Expanse-8B for k values from one to three.}
\end{figure}

\begin{figure}[ht]
\centering
\includegraphics[width=0.8\textwidth]{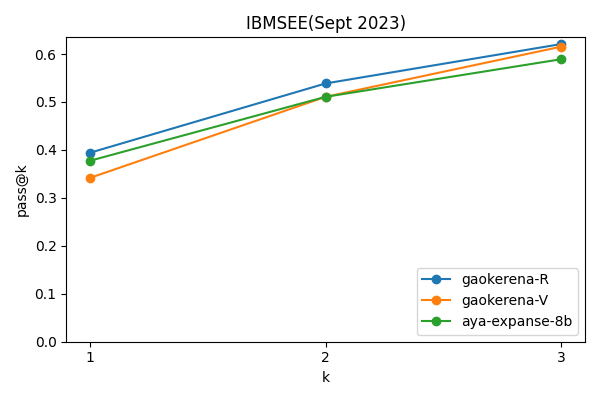}
\caption{Pass@k on the September 2023 KOPP session.}\label{fig9}
\Description{Line plot showing pass@k scores for Gaokerena-R, Gaokerena-V, and Aya-Expanse-8B for k values from one to three.}
\end{figure}

We also report pass@k~\cite{b37}, the probability that at least one of $k$ sampled answers is correct. With $M=5$ samples and $C$ correct samples, pass@k is computed as
\begin{equation}
\mathrm{pass@k}=1-\frac{\binom{M-C}{k}}{\binom{M}{k}}.
\label{fr:passatk}
\end{equation}
We average pass@k for $k\leq3$ over all questions; $k=4$ is omitted because a system could trivially cover all four options. In Figs.~\ref{fig8} and~\ref{fig9}, the Gaokerena-R curve is above those of Gaokerena-V and the backbone in nearly all categories, with anatomy an exception. This suggests that Gaokerena-R's correct options are more often present among its first few samples, but pass@k differences were not tested for significance and should be read alongside Table~\ref{tab:cotsc_sig}. These curves do not measure calibrated confidence. Gaokerena-V's response variability may improve pass@k at higher $k$, but that possibility was not tested.

\subsection{English--Persian Performance Gap}\label{subsec6.3}

Because FA\_MED\_MMLU is translated from English MMLU items, the corresponding English questions can be used to compare language conditions on the same underlying questions. Aya-Expanse-8B was evaluated on the original English questions under direct prompting and under CoT self-consistency with five generations and the three-vote rule. English prompts were translated from the Persian prompts and instructed the model to answer in English. Where experiment files included more than one generation per question, the first was used; CoT-SC uses five generations per question. These are single-run comparisons, and exact prompt, translation, extraction, and contamination equivalence across languages is not established.

Under direct prompting, Aya-Expanse-8B obtains 59.04\% in English (643/1,089; 95\% Wilson interval 56.10--61.93\%) compared with 46.65\% in Persian, a difference of 12.40 points under the reported unpaired test ($p<0.001$). Under CoT-SC, the English score is 63.18\% (688/1,089; interval 60.27--65.99\%; 88.9\% of questions reach a three-vote majority), versus 46.92\% in Persian, a difference of 16.25 points ($p<0.001$). CoT-SC improves the English result by 4.13 points over direct prompting ($p=0.048$ under the unpaired test), while the Persian backbone changes by 0.27 points. Paired tests on per-question predictions would be preferable for formal inference.

\begin{table}[ht]
\caption{Pooled accuracy on the six medical MMLU subjects (1,089 questions) in English and Persian. $\Delta$ is English Aya-Expanse-8B minus the Persian system in percentage points. Reported $p$-values use unpaired two-proportion $z$-tests and are uncorrected. The verifier configuration has no English counterpart; its two differences compare it with English direct prompting and English CoT-SC, respectively.}\label{tab:en_fa}
\small
\setlength{\tabcolsep}{4pt}
\begin{tabular}{@{}llccc@{}}
\toprule
Prompting & Model & Accuracy & $\Delta$ vs. English & $p$ \\
\midrule
Straight & English Aya-Expanse-8B & 59.04 & --- & --- \\
 & Persian Aya-Expanse-8B & 46.65 & +12.40 & $<0.001$ \\
 & Persian Gaokerena-V & 49.31 & +9.73 & $<0.001$ \\
 & Persian Gaokerena-R & 46.10 & +12.95 & $<0.001$ \\
\midrule
CoT-SC, no verifier & English Aya-Expanse-8B & 63.18 & --- & --- \\
 & Persian Aya-Expanse-8B & 46.92 & +16.25 & $<0.001$ \\
 & Persian Gaokerena-V & 40.13 & +23.05 & $<0.001$ \\
 & Persian Gaokerena-R & 48.58 & +14.60 & $<0.001$ \\
\midrule
CoT-SC + verifier & Persian Gaokerena-R & 52.80 & +6.24 / +10.38 & 0.003 / $<0.001$ \\
\bottomrule
\end{tabular}
\end{table}

Relative to the English reference, Gaokerena-V narrows the direct-prompting gap from 12.40 to 9.73 points in the observed results. Gaokerena-R with the verifier is 6.24 points below English direct prompting and 10.38 points below English CoT-SC. These are point-estimate comparisons rather than evidence of a causal effect from training. Gaokerena-V's direct-prompting improvement over the backbone is not significant ($p=0.21$), and Gaokerena-R's verifier configuration uses substantially more inference computation. Under matched CoT-SC without a verifier, the gap is 14.60 points for Gaokerena-R and 16.25 for the backbone, a narrowing of 1.65 points. The paired exact McNemar test for Gaokerena-R versus the backbone is not significant ($p=0.190$). Gaokerena-V widens the gap under CoT-SC to 23.05 points.

\begin{table}[ht]
\caption{Direct-prompting accuracy by subject in English and Persian. Gap columns are English Aya-Expanse-8B minus the indicated Persian system, in percentage points. Values are descriptive single-run results.}\label{tab:en_fa_subject}
\small
\begin{tabular}{@{}lrrrrr@{}}
\toprule
Subject & $n$ & English Aya & Persian Aya & Persian Gaokerena-V & Gap (Aya / V) \\
\midrule
Anatomy & 135 & 59.26 & 40.74 & 48.15 & 18.52 / 11.11 \\
Medical genetics & 100 & 64.00 & 49.00 & 53.00 & 15.00 / 11.00 \\
College medicine & 173 & 53.18 & 44.51 & 43.93 & 8.67 / 9.25 \\
Clinical knowledge & 265 & 65.28 & 52.08 & 55.47 & 13.20 / 9.81 \\
Professional medicine & 272 & 47.79 & 45.59 & 47.06 & 2.20 / 0.73 \\
College biology & 144 & 72.22 & 45.14 & 47.22 & 27.08 / 25.00 \\
\midrule
Pooled & 1,089 & 59.04 & 46.65 & 49.31 & 12.40 / 9.73 \\
\bottomrule
\end{tabular}
\end{table}

The largest direct-prompting gaps occur in college biology (27.08 points for the backbone) and anatomy (18.52 points); the gap is small in professional medicine (2.20 points), where English accuracy is itself relatively low. Gaokerena-V narrows the gap most in anatomy and clinical knowledge. Subject subsets contain 100--272 questions, so these differences are descriptive rather than tested.

A plausible explanation for the observed language gap is the smaller quantity of Persian medical data available to multilingual models, but this study does not isolate that mechanism. English questions may be better represented in training data; translation artefacts may affect Persian question difficulty; the result may partly reflect general Persian proficiency, tokenization, or answer-format adherence rather than medical knowledge alone; and prompts, decoding, extraction, and generation selection were not fully identical across languages. As with the other results, single-run unpaired tests do not quantify seed variability or use the pairing among questions. The experiments establish a substantial English--Persian performance gap under the tested conditions, but do not establish what fraction of it is caused by data scarcity or what fraction is recovered by the training methods.

\paragraph{Why not translate at inference time?} A possible alternative is to translate Persian questions into English, answer them with an English medical model, and translate the response back into Persian. We did not evaluate this pipeline, and stronger translators than the small on-device systems tried in~\cite{b1} may make it competitive. Direct Persian models may still be useful where local deployment, latency, privacy, or avoidance of compounded translation errors matters. They may also reduce reliance on resources designed primarily for another language. The training work reported here used modest compute (about 19 GPU-hours on one A100 for Gaokerena-V~\cite{b1} and about one GPU-hour on one H100 for the DPO stage of Gaokerena-R, excluding data generation~\cite{b2}), but the model results do not show that direct Persian models can replace English systems or translation pipelines. We present direct Persian modeling as a direction worth pursuing, not a demonstrated replacement for translation.

\section{Limitations}\label{sec:limitations}

\begin{itemize}
\item \textbf{Resource representativeness and quality.} The corpus is dominated by a small number of websites and includes syndicated, promotional, and loosely medical material. MF3QA is strongly selected by answer length, has source-confounded splits, contains repetitive answers, and includes sensitive medical content. Its test distribution differs from training. A small number of retained pairs (48) could not be linked to the released raw crawl. KOPP includes PDF-extraction artefacts and answer-key imbalance; FA\_MED\_MMLU was translated but not audited through back-translation or a second annotator.
\item \textbf{Privacy, permissions, and release conditions.} Automated screens identify some residual contact details and references to physicians, but they do not establish that all identifying or sensitive information has been removed. Public accessibility does not itself establish permission for redistribution or all downstream uses. The resources should be reused in accordance with applicable source terms, dataset/model licenses, and any necessary privacy or access controls. This analysis is not a legal determination or a guarantee of anonymization.
\item \textbf{Evaluation scope.} Results are single runs and quantify sampling variability over questions, not over training or decoding seeds. Main comparisons use unpaired tests despite shared question sets; paired per-question tests and confidence intervals for paired differences are preferable. KOPP evaluation covers only one 168-question session, and pooled MMLU results are weighted toward two subjects. Gaokerena-R's 52.80\% configuration uses CoT self-consistency and a verifier and is not compute-matched to the direct baselines. The backbone was not tested through the same pipeline. Under matched CoT-SC without a verifier, Gaokerena-R's advantage over the backbone is not significant, and unanswered questions count as incorrect without separate coverage reporting.
\item \textbf{Training effects.} The experiments do not isolate corpus adaptation, MF3QA instruction tuning, the two preference-generation strategies, or DPO through ablations. The observed results therefore do not identify the independent contribution of each resource or training stage. Training data and preference pairs have not been comprehensively checked for contamination against evaluation sets.
\item \textbf{English--Persian comparison.} The observed gap does not isolate Persian data scarcity from benchmark contamination, translation artefacts, general Persian proficiency, answer extraction, or protocol differences. English and Persian runs were evaluated once, and stronger controls would require paired analysis and better-matched prompts and decoding.
\item \textbf{Free-form evaluation and baselines.} We do not report a validated free-form evaluation of Gaokerena-V on MF3QA. Corpus overlap also makes the development split unsuitable as an independent evaluation without filtering. Baselines are limited to three general-purpose models of similar size; newer multilingual models and larger systems were not tested.
\item \textbf{Intended use.} The models are research artefacts, not clinically validated systems, and should not be used to provide medical advice, particularly medication-dosing advice. MCQ accuracy is a limited proxy for medical knowledge and does not establish clinical safety or competence.
\end{itemize}

\section{Ethical Considerations}\label{sec:ethics}

This study involved no direct interaction with human participants. The text resources were collected from publicly accessible Persian medical websites and forums; no private or access-restricted pages were intentionally collected. Public availability does not, on its own, resolve questions of consent, privacy, copyright, website terms, or redistribution rights. The article reports automated screening for contact information and other indicators, but these heuristics can miss identifying details and can produce false positives. The cleaned MF3QA release may still include references to physicians and sensitive medical topics, including questions that appear to come from minors. These characteristics should not be interpreted as proof that the data are fully anonymized or suitable for every downstream use.

Before continued distribution or reuse, the relevant source terms and licenses should be checked for each resource, and residual privacy risks should be reviewed against the intended release conditions. Any additional filtering, masking, access restriction, or record removal should be documented and applied to the distributed artefacts where needed, rather than left solely to downstream users. The licenses and use conditions of both derived models should be stated in the release documentation, taking account of the terms applicable to the Aya-Expanse-8B backbone. Gaokerena-V and Gaokerena-R are research models and are not validated for clinical use.

\section{Conclusion}\label{sec:conclusion}

This article presents and characterizes a suite of Persian medical language resources: an 85.5-million-word web corpus, MF3QA patient--doctor QA pairs, an expanded collection of Iranian medical-exam questions, translated medical MMLU questions, and a machine-translated MCQ training set. The analyses document scale and topical coverage as well as source imbalance, orthographic variation, redundancy, selection effects, differences between QA splits, benchmark artefacts, and privacy-related indicators. These characterizations are a central contribution because they make visible the conditions under which the resources can be used and the limitations that future users should account for.

Gaokerena-V and Gaokerena-R provide an application study of the resources. Under direct prompting, Gaokerena-V reaches 49.31\% on translated medical MMLU, 2.66 percentage points above Aya-Expanse-8B, but the difference is not statistically significant under the reported test ($p=0.21$). Gaokerena-R reaches 52.80\% with five-sample self-consistency and a verifier, but this higher-cost configuration is not compute-matched to the directly prompted backbone. Under matched CoT self-consistency without a verifier, Gaokerena-R scores 48.58\% versus 46.92\% for the backbone; the paired exact McNemar test is not significant ($p=0.190$; unpaired $p=0.44$). The English comparison also shows a substantial performance gap for the corresponding questions, but the experiments do not isolate training effects from translation, benchmark exposure, or inference-time computation.

The evidence therefore supports the value of assembling and systematically characterizing Persian medical resources, while the model results should be read as protocol-specific demonstrations rather than proof of a general training advantage or clinical readiness. Priorities for further work include paired per-question analyses, repeated runs, contamination checks, fuller audits of translation and source quality, training ablations, validated free-form evaluation, and appropriate privacy and licensing review for all distributed artefacts.

\section*{Data and Code Availability}

The medical corpus, MF3QA and its uncleaned crawl, the KOPP and FA\_MED\_MMLU benchmarks, and the translated MedMCQA training set are hosted at \url{https://huggingface.co/datasets/gaokerena} (the translated MedMCQA set is in the Gaokerena-R repository). Code, model weights, and generated responses are available through \url{https://github.com/Mehrdadghassabi/Gaokerena-V} and \url{https://github.com/Mehrdadghassabi/Gaokerena-R}. Availability does not imply unrestricted reuse: each artefact should be used according to its applicable license, source terms, and stated privacy and safety limitations. The release documentation should identify these conditions for each dataset and model.

\section*{Author Contributions}
M.G. led and carried out the core research. H.B.K. supervised the research. Z.K. and M.T. were responsible for data collection. A.P. was responsible for translation. P.R. and S.H. contributed to technical aspects of the work. A.E. contributed to the technical aspects of the present extended journal study. A.E. was not an author of the two earlier conference papers cited in this article; her contribution is to this expanded manuscript. All authors read and approved the final manuscript.

\section*{Competing Interests}
The authors declare no competing interests.

\begin{acks}
This research received no external funding and was self-funded by the authors.
\end{acks}


\end{document}